\documentclass[times,twocolumn,final]{elsarticle}

\usepackage{medima}
\usepackage{framed,multirow}

\usepackage{amssymb}
\usepackage{latexsym}
\usepackage{graphicx}
\usepackage{makecell}
\usepackage{array}
\usepackage{mdwmath}
\usepackage{mdwtab}
\usepackage{eqparbox}
\usepackage{url}
\usepackage{multirow}
\usepackage{algorithm}
\usepackage{url}
\usepackage{xcolor}
\usepackage{multirow}
\usepackage{hyperref}

\definecolor{newcolor}{rgb}{.8,.349,.1}

\usepackage{multirow}
\usepackage{pifont}
\usepackage{amsmath}
\usepackage{CJKutf8}
\usepackage{algpseudocode} 
\usepackage{lineno}
\usepackage[normalem]{ulem} 

\definecolor{r}{RGB}{255,0,0}
\definecolor{b}{RGB}{0,0,255}

\journal{Medical Image Analysis}

\begin{document}

\verso{Yonghuang Wu \textit{et~al.}}

\begin{frontmatter}

\title{SAMPO-Path: Segmentation Intent-Aligned Preference Optimization for Pathology Foundation Model Segmentation}

\author[1]{Yonghuang \snm{Wu}}
\author[1]{Wenwen \snm{Zeng}}
\author[1]{Pengfei \snm{Song}}
\author[1]{Chengqian \snm{Zhao}}
\author[1]{Xuan \snm{Xie}}
\author[1]{Guoqing \snm{Wu}}
\author[1]{Jinhua \snm{Yu}\corref{cor1}}
\cortext[cor1]{Corresponding author. Email: jhyu@fudan.edu.cn}

\address[1]{School of Biomedical Engineering and Technological Innovation, Fudan university, Shanghai, China.}

\received{}
\finalform{}
\accepted{}
\availableonline{}

\begin{abstract}
Foundation models have recently demonstrated strong capabilities in multi-object segmentation via visual prompts. However, the high cellular density and pronounced heterogeneity of histopathology images pose persistent challenges. Current fine-tuning paradigms, predominantly relying on pixel-level supervision, fail to capture the granularity of clinical segmentation intent (e.g., discriminatively segmenting all nuclei of a specific type). In practice, such intents are often expressed through diverse visual prompts of varying quality. Consequently, current models struggle with prompt-intent alignment, where slight variations in prompts lead to inconsistent outputs. Such misalignment constrains semantic understanding and hinders the clinical reliability of foundation models. 

We present SAMPO (Segmentation Anything Model with Preference Optimization), a preference-aligned fine-tuning framework designed to explicitly align pathology foundation models with clinical segmentation intents. 
In SAMPO, we pioneer the adaptation of Direct Preference Optimization (DPO) to pure vision foundation models, where preference-driven learning empowers accurate visual predictions from minimal prompts. 
The framework introduces three key innovations: 
(1) Online prompt-centric preference mining, which automatically constructs preference pairs for a given intent by synthesizing prompts of varying quality; 
(2) Multi-mask preference learning, which exploits the intrinsic ambiguity in model outputs to provide fine-grained ranking feedback; and 
(3) a hybrid loss that couples preference optimization with pixel-level supervision to ensure training stability. 
Extensive experiments on multiple pathology benchmarks demonstrate that SAMPO consistently improves both segmentation accuracy and adherence to clinical intent, 
exhibiting superior robustness to prompt variations and yielding consistent outputs even in dense, heterogeneous histopathology scenes.
\end{abstract}


\end{frontmatter}



\section{Introduction}

Recent advances in promptable segmentation models such as SAM \cite{kirillov2023segany} have enabled unprecedented flexibility in adapting to diverse visual queries. By conditioning mask generation on user-specified prompts (e.g., points, boxes, or texts), these models exhibit strong zero-shot generalization across objects and domains. However, despite their impressive scalability, current fine-tuning paradigms remain pixel-centric—they minimize low-level reconstruction errors rather than capturing the semantic intention embedded in user prompts. As a result, models often produce inconsistent or ambiguous masks when given imprecise or semantically complex prompts, limiting their utility in real-world, interactive segmentation. 
This limitation becomes particularly pronounced in digital pathology. In routine clinical workflows, pathologists are rarely interested in isolating a single object instance (e.g., segmenting one nucleus or all pixels within a bounding box). Instead, they frequently require the quantitative assessment of specific cell populations—such as tumor cells, lymphocytes, stromal nuclei, or microvessels—where analyzing spatial distributions, densities, and morphological statistics is fundamental for diagnosis, grading, and prognosis. As a result, segmentation systems that fail to align with such high-level clinical intent may produce masks that are technically accurate yet clinically suboptimal.

\begin{figure}
    \centering
    \includegraphics[width=1\linewidth]{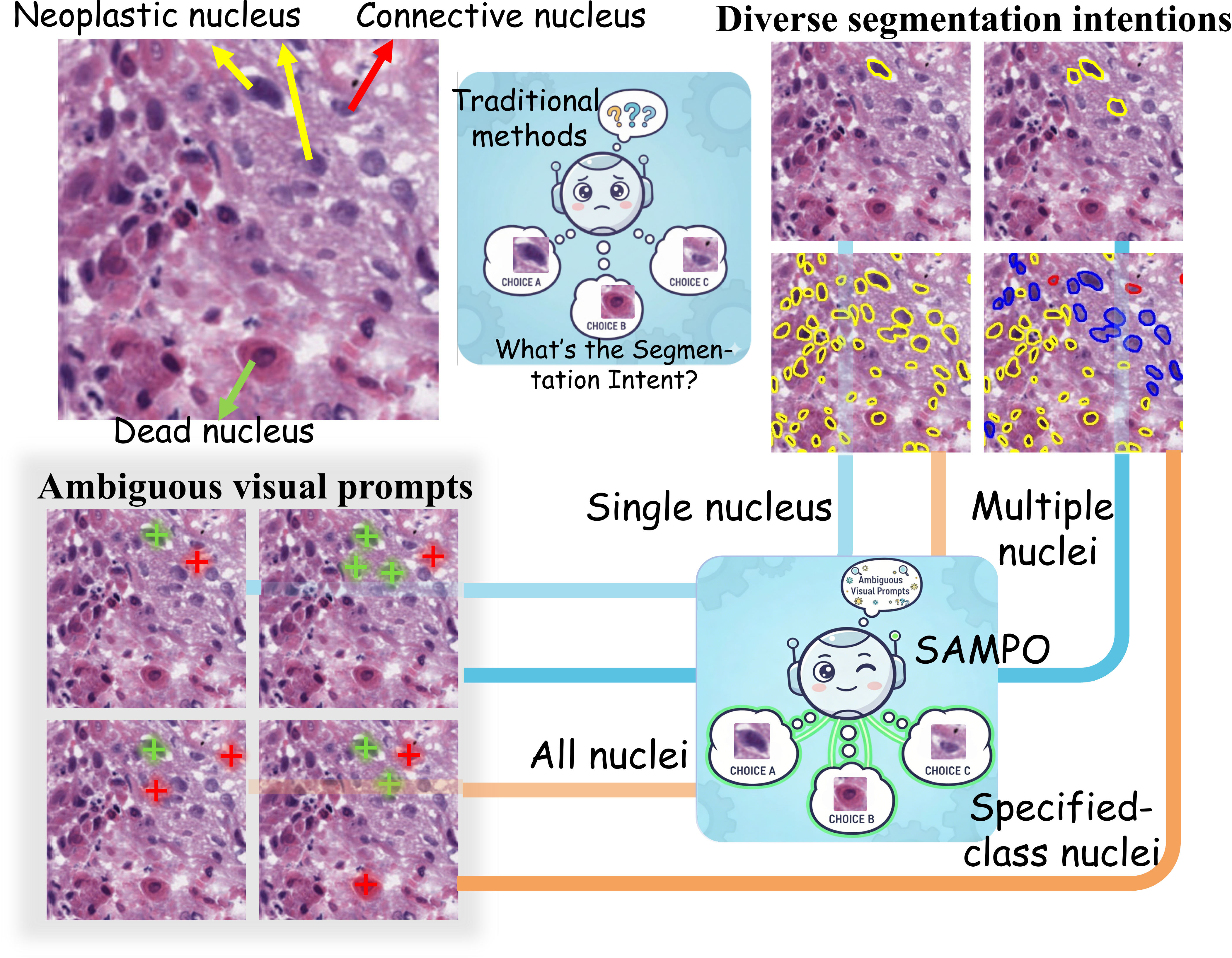}
    \caption{Illustration of preference-optimized segmentation versus traditional pixel-centric paradigms.}
    \label{fig:inno}
\end{figure}

This gap reflects a broader challenge: attaining preference alignment, ensuring that model outputs are consistent with human expectations and task intent, has become a central objective in large language models, yet remains underexplored in promptable visual systems. 
We contend that attaining preference alignment, which means ensuring the model responds in a manner consistent with human expectations, is crucial for the next generation of promptable vision systems. In contrast to conventional pixel-level objectives, preference alignment explicitly optimizes for the relative quality of model outputs conditioned on prompt variations. In the context of segmentation, this involves learning to prefer masks that better reflect the user’s intended object, rather than those that merely maximize overlap scores under a fixed annotation.

To realize this idea, we introduce SAMPO, a framework for preference-aligned fine-tuning of promptable segmentation models (shown in Figure~\ref{fig:inno}). SAMPO adapts the principles of preference optimization to the visual domain while addressing its unique challenges—dense output spaces, structured ambiguities, and the need for stable optimization. The framework is built on three core innovations:
\begin{itemize} 
    \item Online prompt-centric preference mining. Instead of relying on static preference datasets or manual annotations, SAMPO dynamically constructs preference pairs during training by analyzing how different prompt variations affect segmentation quality relative to ground-truth objects.
    \item Fine-grained learning via multi-mask ambiguity. Modern segmentation foundation models produce multiple candidate masks per prompt, reflecting uncertainty and object hierarchies. SAMPO leverages these multi-mask outputs as internal preference signals, guiding the model to rank and refine its own hypotheses.
    \item Hybrid optimization for stable alignment. Direct preference learning in segmentation can be unstable due to the high-dimensional mask space. SAMPO stabilizes training with a hybrid objective that combines preference alignment with pixel-level supervision, ensuring both perceptual consistency and quantitative fidelity.
\end{itemize}

Through these designs, SAMPO bridges the gap between prompt sensitivity and segmentation reliability, providing a principled way to align model behavior with prompt semantics. Experiments on multiple promptable segmentation benchmarks demonstrate that SAMPO consistently improves accuracy, robustness, and responsiveness to user intent—advancing the field toward intention-aligned visual foundation models.

\section{Related Work}

\subsection{Medical Visual Foundation Model}

In recent years, the emergence of Medical Visual Foundation Models (VFMs), particularly the SAM, has marked a shift in the image segmentation domain from task-specific models to general models leveraging large-scale pre-training \cite{kirillov2023segany}. SAM has demonstrated remarkable zero-shot segmentation capabilities on natural images; however, when directly applied to medical images, especially in dense instance segmentation tasks such as nuclei and organ segmentation \cite{10822028, na2024segment, instasam}, it still faces several challenges.

\textbf{Domain Gap}: SAM is primarily trained on natural images, resulting in decreased performance when handling medical images characterized by different features, such as low contrast, complex textures, and ambiguous boundaries \cite{ma2024segment}. To address this issue, researchers have adapted SAM by fine-tuning it on large medical datasets \cite{cheng2024unleashing}. However, such full-scale fine-tuning or large-scale data collection approaches are computationally expensive and difficult to cover all medical subfields \cite{wang2024promise}.

\textbf{Prompt Efficiency Bottleneck}: SAM's interactive design relies on provided prompts (e.g., points or boxes) to identify targets \cite{kirillov2023segany}. For images containing dozens of nuclei, providing precise dense prompts one by one is impractical, significantly increasing annotation cost and time \cite{liu2024rethinking}. Studies have shown that simply increasing the number of sparse prompts does not guarantee improved segmentation quality and may even lead to performance degradation \cite{na2024segment}. To overcome this bottleneck, some works have explored parameter-efficient fine-tuning \cite{hu2022lora} and automatic prompt generation. For instance, Segment Any Cell \cite{na2024segment} employs an auxiliary network to automatically generate dense point prompts and then utilizes low-rank adaptation for fine-tuning. Although this approach reduces human interaction, it fundamentally compensates for the model's inadequate understanding of sparse instructions by generating a large number of prompts. Existing works \cite{chen2025sam, 10822028} have primarily focused on automating prompt generation without addressing a more fundamental issue: how to enable the model to better understand the user intent underlying a small number of prompts.

\subsection{Preference Learning for Model Alignment}

Preference learning, particularly through Reinforcement Learning from Human Feedback (RLHF) \cite{ouyang2022training}, has become a central technique for aligning large language models (LLMs) \cite{radford2019language} with complex human intentions \cite{jiang2024survey}. This paradigm has achieved significant success in enhancing the instruction-following capabilities and naturalness of interaction in LLMs \cite{rafailov2023direct, deepseekai2025deepseekr1incentivizingreasoningcapability}.

Recent advances in multimodal large models \cite{liu2023llava, bai2025qwen2} have demonstrated significant progress in text-based reasoning. Several works further extend these capabilities to visual grounding tasks \cite{zhu2023minigpt, lai2023lisa, yang2023lisaPP, huang2025medseg, tong2025medisee}, with emerging efforts on preference learning for vision-enabled models \cite{huang2025sam, wang2025pixelthink, liu2025visionreasoner, liu2025seg, pan2025dino}. However, a critical limitation persists: existing approaches predominantly optimize the large language model components (with the exception of \cite{pan2025dino}), while largely neglecting optimization of visual foundation models (VFMs) – or even bypassing them entirely.

This oversight is particularly consequential given the operational efficiency advantage of VFMs: sparse visual prompts incur substantially lower deployment costs compared to multimodal prompt engineering. We thus identify a pivotal research gap: The potential of sparse prompting in VFMs remains fundamentally constrained by two unresolved tensions:
(1) Cost-Precision Mismatch: The absence of intent-aware alignment mechanisms prevents cost efficiency from translating to high-fidelity dense predictions.
(2) Intent-Response Discrepancy: Implicit semantic demands in sparse prompts cannot be autonomously interpreted by current visual models.

\section{Methodology}

\begin{figure*}[h]
    \centering
    \includegraphics[width=1\linewidth]{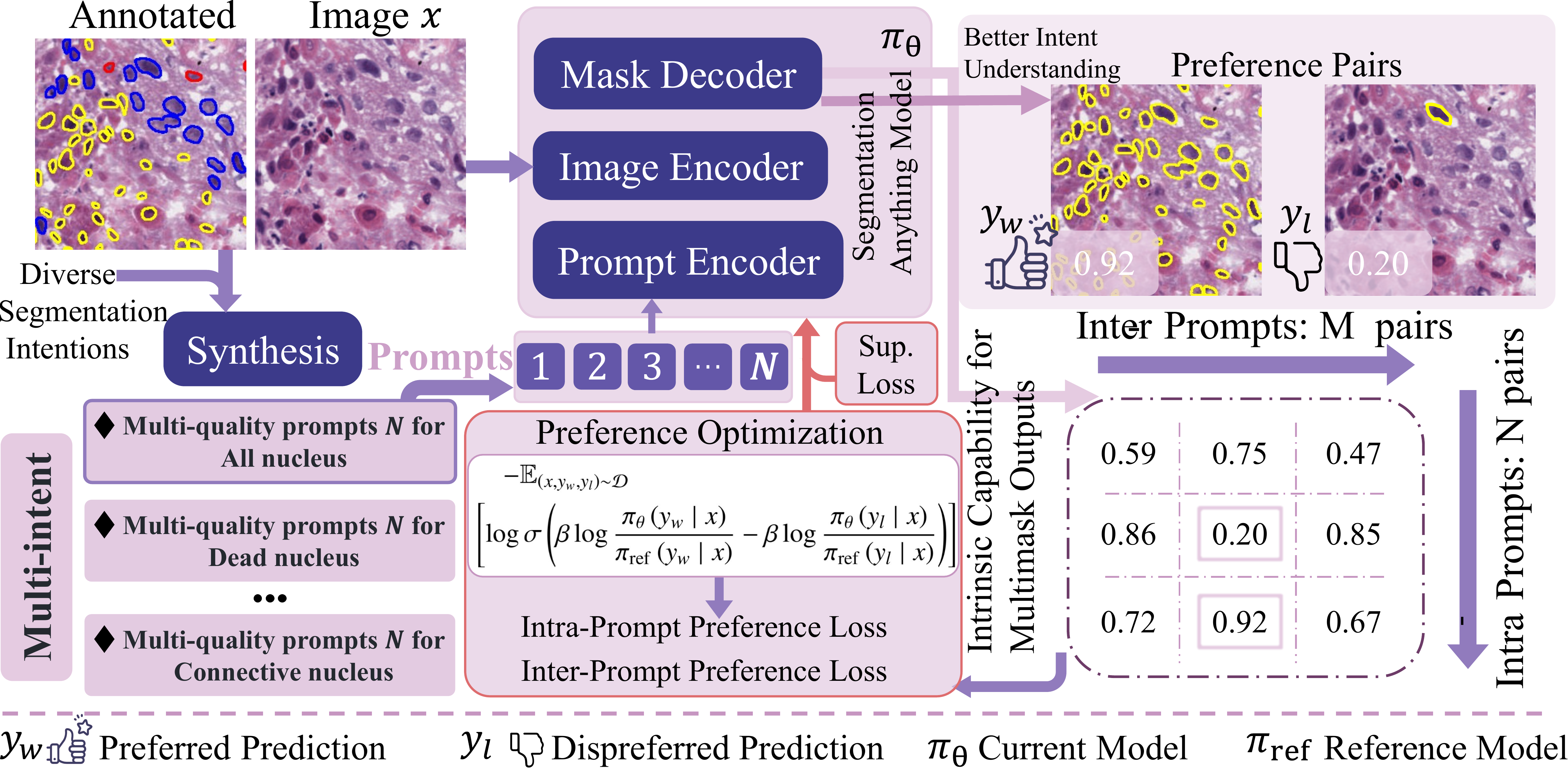}
    \caption{Overview.}
    \label{fig:method}
\end{figure*}

\subsection{Preliminaries and Motivation}

\subsubsection{Clinical Segmentation Intent in Pathology}
In clinical histopathology, segmentation is not a singular, unambiguous task. The clinician's intent—whether to segment a specific nucleus of interest, all nuclei of a particular phenotypic class, or every nucleus in a field of view—defines the semantic goal. A critical challenge is that this \textit{fixed intent} can be communicated to a promptable foundation model through a spectrum of visual prompts (e.g., point clicks) of varying precision and informativeness. Current fine-tuning paradigms, which minimize pixel-level errors like Binary Cross-Entropy (BCE) loss, fail to model this relationship. They treat all prompts leading to a reasonable mask as equally good, lacking a mechanism to teach the model to \textbf{discriminate and prefer prompts that better encapsulate the clinical intent}. This leads to suboptimal performance when prompt quality is imperfect—a frequent scenario in practice.

\subsubsection{Preference Optimization for Intent Alignment}
We posit that aligning a segmentation model with clinical intent is analogous to aligning a language model with human preferences. The core problem shifts from "Is this mask correct?" to "Given this clinical intent, does this mask (from prompt A) better fulfill it than that mask (from prompt B)?". This is a \textbf{relative ranking} problem, for which Direct Preference Optimization (DPO) offers a principled and lightweight solution \cite{rafailov2023direct}.
Given a preference dataset $\mathcal{D}=\left\{\left(x, y_w, y_l\right)\right\}$ where $y_w$ is preferred over $y_l$ for context $x$, DPO optimizes a policy $\pi_\theta$ against a reference policy $\pi_{\mathrm{ref}}$:
\begin{equation}
\label{eq:dpo_original}
\begin{aligned}
\mathcal{L}_{\mathrm{DPO}} & =-\mathbb{E}_{\left(x, y_w, y_l\right) \sim \mathcal{D}} \\
& \left[\log \sigma\left(\beta \log \frac{\pi_\theta\left(y_w \mid x\right)}{\pi_{\mathrm{ref}}\left(y_w \mid x\right)}-\beta \log \frac{\pi_\theta\left(y_l \mid x\right)}{\pi_{\mathrm{ref}}\left(y_l \mid x\right)}\right)\right],
\end{aligned}
\end{equation}
where $\sigma$ is the sigmoid function and $\beta$ a scaling parameter. This loss increases the likelihood of $y_w$ relative to $y_l$, regularized by $\pi_{\mathrm{ref}}$ to prevent drift. We adapt this framework to visual segmentation, where $x$ is an (image, prompt) pair, and preferences are derived from how well the output masks align with the known clinical intent, as proxied by ground-truth masks.

\subsection{The SAMPO Framework}
SAMPO is designed to fine-tune promptable pathology segmentation models (e.g., SAM \cite{kirillov2023segany}) by aligning them with clinical intent through preference optimization. The framework operates on a core principle: for a single training image and a fixed clinical intent related to a specific object (e.g., "segment all neoplastic nuclei"), we can simulate multiple prompting scenarios of varying quality. The resulting masks are ranked by their alignment with the intent (measured via IoU with ground truth), creating the preference pairs needed for DPO-based learning.

\subsubsection{Online Prompt-Centric Preference Mining}
A cornerstone of SAMPO is the online generation of preference data, eliminating the need for pre-collected human rankings. This process is detailed in Algorithm \ref{algo}.

Given an image $I$ and a target object instance with ground-truth mask $O_{\text{gt}}$, we first define a fixed clinical segmentation intent (e.g., "segment this object"). To simulate variable clinical prompting, we synthesize $N$ distinct prompt sets $\mathcal{S} = \{S_1, S_2, ..., S_N\}$. Each $S_k$ represents a valid but quality-varied expression of the \textbf{same intent}, achieved by varying factors such as: the number of positive/negative points, the proximity of points to the object boundary, or the inclusion of ambiguous points. This creates a natural quality gradient among prompts.

The segmentation model $\pi_\theta$ processes each $(I, S_k)$ pair to produce a candidate mask $y_k$. We then compute the alignment score $m_k = \text{IoU}(y_k, O_{\text{gt}})$. A fundamental premise is that prompts of higher quality for the given intent will, on average, yield masks with higher IoU. From the set of candidate masks $\mathcal{Y} = \{y_k\}_{k=1}^N$, we construct preference pairs $(y_w, y_l)$ where $m_w > m_l$. This direct pairing teaches the model that the prompt leading to $y_w$ is a better expression of the clinical intent than the prompt leading to $y_l$.

\begin{algorithm}[t]
\caption{Online Preference Mining for a Fixed Clinical Intent}
\label{algo}
\begin{algorithmic}[1]
\Require Image $I$, Ground Truth Object Mask $O_{\text{GT}}$, Fixed Segmentation Intent
\Ensure Preference Pair Set $\mathcal{P}_{\text{pref}}$
\State $\mathcal{P}_{\text{pref}} \leftarrow \emptyset$
\State Synthesize $N$ prompt sets $\mathcal{S} = \{S_1, \dots, S_N\}$ reflecting the fixed intent with varying quality.
\State $\mathcal{Y} \leftarrow \emptyset$, $\mathcal{M} \leftarrow \emptyset$
\For{$k = 1$ to $N$}
    \State $y_k \leftarrow \pi_\theta(I, S_k)$ \Comment{Generate candidate mask}
    \State $\mathcal{Y} \leftarrow \mathcal{Y} \cup \{y_k\}$
    \State $m_k \leftarrow \text{IoU}(y_k, O_{\text{GT}})$
    \State $\mathcal{M} \leftarrow \mathcal{M} \cup \{m_k\}$
\EndFor
\State Identify index pairs $(i, j)$ where $m_i > m_j$
\For{each qualifying pair $(i, j)$}
    \State $\mathcal{P}_{\text{pref}} \leftarrow \mathcal{P}_{\text{pref}} \cup \{(I, S_i, y_i, S_j, y_j)\}$ \Comment{$(y_i, y_j)$ is a preference pair}
\EndFor
\State \Return $\mathcal{P}_{\text{pref}}$
\end{algorithmic}
\end{algorithm}

\subsubsection{Fine-Grained Learning via Multi-Mask Ambiguity}
SAM-like models natively output multiple candidate masks (e.g., 3) for a single prompt to handle ambiguity. SAMPO repurposes this not as a nuisance, but as a rich source of fine-grained, intra-prompt feedback.
For a single prompt set $S_k$, the model produces $M$ hypotheses $\{y_{k}^{(1)}, y_{k}^{(2)}, ..., y_{k}^{(M)}\}$. Even within the context of a single prompt quality level, these masks exhibit varying alignment with $O_{\text{gt}}$. We evaluate all $M$ masks, rank them by IoU, and construct intra-prompt preference pairs (e.g., $(y_{k}^{(best)}, y_{k}^{(worst)})$).
This mechanism addresses a key pathology challenge: inherent visual ambiguity at object boundaries (e.g., touching nuclei, faint staining). It teaches the model to refine its own ranking of hypotheses, effectively performing self-correction and leading to sharper, more confident boundaries in the final chosen output.

\subsubsection{Preference Optimization Objective for Pathology}
Integrating the inter-prompt (quality-based) and intra-prompt (ambiguity-based) preference mining, we formalize the SAMPO preference loss. Let $\pi_{\text{ref}}$ denote a frozen copy of the initial model.
The Inter-Prompt Preference Loss $\mathcal{L}_{\text{PO1}}$ encourages the model to distinguish between different prompt qualities for the same intent. For each output slot $j$ across the $N$ prompts, we contrast the best and worst masks:
\begin{equation}
\label{eq:inter}
\begin{aligned}
\mathcal{L}_{\text{PO1}} = -\mathbb{E}_{(I, O_{\text{GT}})} \frac{1}{M} \sum_{j=1}^{M} \log \sigma \Bigg( & \log \frac{\pi_\theta(y_{w}^{(j)} \mid I, S_{w})}{\pi_{\text{ref}}(y_{w}^{(j)} \mid I, S_{w})} \\
& - \log \frac{\pi_\theta(y_{l}^{(j)} \mid I, S_{l})}{\pi_{\text{ref}}(y_{l}^{(j)} \mid I, S_{l})} \Bigg),
\end{aligned}
\end{equation}
where $y_{w}^{(j)}$ and $y_{l}^{(j)}$ are the masks from the prompt sets yielding the highest and lowest IoU for the $j$-th output slot, respectively.

The Intra-Prompt Preference Loss $\mathcal{L}_{\text{PO2}}$ focuses on resolving ambiguity within a single prompt:
\begin{equation}
\label{eq:intra}
\begin{aligned}
\mathcal{L}_{\text{PO2}} = -\mathbb{E}_{(I, O_{\text{GT}})} \frac{\lambda}{N} \sum_{k=1}^{N} \log \sigma \Bigg( & \log \frac{\pi_\theta(\hat{y}_w^{(k)} \mid I, S_k)}{\pi_{\text{ref}}(\hat{y}_w^{(k)} \mid I, S_k)} \\
& - \log \frac{\pi_\theta(\hat{y}_l^{(k)} \mid I, S_k)}{\pi_{\text{ref}}(\hat{y}_l^{(k)} \mid I, S_k)} \Bigg),
\end{aligned}
\end{equation}
where $\hat{y}_w^{(k)}$ and $\hat{y}_l^{(k)}$ are the best and worst masks among the $M$ outputs for prompt $S_k$, and $\lambda$ balances the two terms.
The total preference loss is:
\begin{equation}
\label{eq:po_total}
\mathcal{L}_{\text{PO}} = \mathcal{L}_{\text{PO1}} + \mathcal{L}_{\text{PO2}}.
\end{equation}

\subsubsection{Hybrid Optimization for Clinical Stability}
Applying $\mathcal{L}_{\text{PO}}$ alone in the high-dimensional, continuous space of segmentation can lead to instability, where the model satisfies the ranking objective with degenerate or unrealistic masks. This is particularly risky in medical imaging where anatomical plausibility is paramount.
Following best practices from RLHF \cite{rafailov2023direct}, we anchor the preference learning with a strong pixel-level supervisory signal:
\begin{equation}
\label{eq:sup_loss}
\mathcal{L}_{\text{SUP}} = \mathcal{L}_{\text{BCE}}(y_w, O_{\text{GT}}) + \mathcal{L}_{\text{BCE}}(y_l, O_{\text{GT}}).
\end{equation}
Crucially, supervision is applied to \textbf{both} the preferred and dispreferred masks in a pair. This ensures both outputs remain within the manifold of valid segmentations, preventing collapse and providing a stable learning signal. It also guarantees that even responses to lower-quality prompts are nudged towards correctness.

The final SAMPO training objective is a weighted hybrid:
\begin{equation}
\label{eq:total_loss}
\mathcal{L}_{\text{SAMPO}} = \mathcal{L}_{\text{SUP}} + \alpha \cdot \mathcal{L}_{\text{PO}},
\end{equation}
where $\alpha$ controls the strength of preference alignment (default: 1.0). This combination enables SAMPO to simultaneously achieve high pixel-level accuracy and superior alignment with the nuanced semantics of clinical segmentation intent, making it uniquely suited for advancing pathology foundation models.

\section{Experiments}

\subsection{Dataset and Evaluation}

\begin{table*}[!ht]
\centering
\caption{Comparison results of nuclei segmentation on two tasks of PanNuke and CoNSeP. Metric: IoU and Dice Score. The best results are highlighted in \textbf{bold}.}
\label{tab:pncn}
\setlength{\tabcolsep}{3pt} 
\begin{tabular}{c|c|cc|cc|cc|cc|cc|cc|cc}
\hline
\multirow{2}{*}{}           & \multirow{2}{*}{Ratio} & \multicolumn{2}{c|}{U-Net}      & \multicolumn{2}{c|}{SwinUNet}   & \multicolumn{2}{c|}{H-SAM}      & \multicolumn{2}{c|}{SAN}        & \multicolumn{2}{c|}{InstaSAM}   & \multicolumn{2}{c|}{MedSAM}     & \multicolumn{2}{c}{Ours}        \\ \cline{3-16} 
                            &                        & \multicolumn{1}{c}{IoU} & Dice  & \multicolumn{1}{c}{IoU} & Dice  & \multicolumn{1}{c}{IoU} & Dice  & \multicolumn{1}{c}{IoU} & Dice  & \multicolumn{1}{c}{IoU} & Dice  & \multicolumn{1}{c}{IoU} & Dice  & \multicolumn{1}{c}{IoU} & Dice  \\ \hline
\multirow{4}{*}{PanNuke-T1} & 10\%     & 55.49     & 71.37     & 52.97     & 69.26     & 48.04     & 64.90     & 51.25     & 67.77     & 59.58     & 74.67     & 64.51     & 78.43     & \textbf{68.28}     & \textbf{81.15} \\ 
                            & 20\%     & 57.36     & 72.90     & 56.43     & 72.15     & 50.14     & 66.79     & 56.91     & 72.54     & 61.45     & 76.12     & 65.25     & 78.97     & \textbf{69.40}     & \textbf{81.94} \\
                            & 30\%     & 58.29     & 73.65     & 56.93     & 72.55     & 51.68     & 68.14     & 59.93     & 74.95     & 62.91     & 77.23     & 67.60     & 80.67     & \textbf{69.31}     & \textbf{81.87} \\
                            & 100\%    & 62.96     & 77.27     & 61.34     & 76.04     & 54.18     & 70.28     & 60.51     & 75.40     & 66.10     & 79.59     & 69.46     & 81.98     & \textbf{70.36}     & \textbf{82.60} \\ \hline
\multirow{4}{*}{PanNuke-T2} & 10\%     & 27.82     & 43.53     & 22.09     & 36.19     & 25.28     & 40.36     & 25.32     & 40.41     & 11.86     & 21.21     & 30.81     & 47.11     & \textbf{50.96}     & \textbf{67.51} \\
                            & 20\%     & 29.01     & 44.97     & 21.80     & 35.80     & 28.18     & 43.97     & 29.35     & 45.38     & 17.16     & 29.29     & 36.49     & 53.47     & \textbf{61.14}     & \textbf{75.88} \\
                            & 30\%     & 29.74     & 45.85     & 22.26     & 36.41     & 30.93     & 47.25     & 31.04     & 47.37     & 22.68     & 36.97     & 39.34     & 56.47     & \textbf{68.58}     & \textbf{81.36} \\
                            & 100\%    & 33.75     & 50.47     & 30.03     & 46.19     & 35.13     & 51.99     & 36.66     & 53.65     & 37.76     & 54.82     & 41.42     & 58.58     & \textbf{69.30}     & \textbf{81.87} \\ \hline
\multirow{4}{*}{CoNSeP-T1}  & 10\%     & 31.39     & 45.85     & 16.72     & 28.28     & -& -& -& -& 16.37     & 27.30     & 33.89     & 47.71     & \textbf{34.09}     & \textbf{49.46} \\
                            & 20\%     & 47.06     & 63.40     & 31.71     & 45.70     & 24.64     & 38.69     & 35.48     & 51.51     & 21.24     & 34.14     & 48.17     & 63.92     & \textbf{49.14}     & \textbf{65.20} \\
                            & 30\%     & 49.92     & 66.33     & 49.38     & 64.89     & 37.33     & 53.85     & 51.40     & 67.49     & 31.44     & 46.33     & 51.92     & 65.88     & \textbf{58.45}     & \textbf{73.25} \\ 
                            & 100\%    & 51.84     & 68.01     & 52.75     & 68.76     & 50.74     & 66.72     & 55.64     & 71.31     & 34.62     & 49.95     & 61.67     & 75.92     & \textbf{64.73}     & \textbf{78.17} \\ \hline
\multirow{4}{*}{CoNSeP-T2}  & 10\%     & 10.82& 17.25& 15.25& 15.53& 18.56     & 29.76& -& -& -& -& 21.23     & 31.35     & \textbf{37.93}     & \textbf{51.92} \\
                            & 20\%     & 12.99& 19.13& 18.16& 18.97& 20.69     & 32.75& 15.70     & 25.12& 18.38     & 29.98& 35.36     & 47.65     & \textbf{39.75}     & \textbf{52.90} \\
                            & 30\%     & 15.04& 22.76& 21.09& 20.98& 22.46     & 34.73& 17.35     & 29.56& 20.45     & 31.98& 35.71     & 48.71     & \textbf{45.27}     & \textbf{57.92} \\
                            & 100\%    & 16.67& 24.86& 22.41     & 22.17     & 25.27     & 38.01& 22.63     & 34.26& 23.96     & 36.33& 40.81     & 54.46     & \textbf{46.69}     & \textbf{61.56} \\ \hline

\end{tabular}
\end{table*}

To comprehensively evaluate SAMPO's ability to align with clinical segmentation intent and its generalizability across diverse histopathology domains, we employ a multi-dataset benchmark. Our experiments are structured into two parts: (1) in-domain evaluation under varying data regimes to assess learning efficiency, and (2) cross-domain zero-shot transfer to evaluate generalization robustness.

\textbf{Training Datasets.} We utilize two datasets for model development and primary evaluation:

\textit{PanNuke} \cite{gamper2020pannuke} contains over 200,000 nuclei from 19 tissue types across 5 centers, annotated with 5 semantic categories (Neoplastic, Inflammatory, Connective, Dead, Epithelial). This diversity makes it an ideal foundation for learning promptable segmentation.

\textit{CoNSeP} \cite{graham2019hover} provides 24,319 nuclei from colorectal adenocarcinoma tissue, annotated with 7 categories (Inflammatory, Epithelial, Spindle-shaped, Miscellaneous, and their malignant variants). Following the original protocol, the annotations are consolidated into 4 semantic categories: Inflammatory, Epithelial, Spindle-Shaped, and Miscellaneous. It serves as a complementary benchmark with finer-grained semantic distinctions.

\textbf{Evaluation Tasks.} We design two core prompting tasks on both PanNuke and CoNSeP:

\textit{Universal Nuclei Segmentation (T1)}: Given sparse point prompts on a few arbitrary nuclei, the model must segment all nuclei in the image. This tests the model's ability to infer a common structural pattern ("nucleus-ness") from minimal cues.

\textit{Category-Specific Nuclei Segmentation (T2)}: Given sparse point prompts on nuclei of a specific semantic type (e.g., Inflammatory), the model must segment all nuclei belonging to that same category. This evaluates a higher level of intent alignment, requiring the model to jointly reason about visual morphology and semantic consistency.

\textbf{In-Domain Evaluation.} To assess data efficiency, we train all models under four data regimes: 10\%, 20\%, 30\%, and 100\% of the training set. We compare SAMPO against representative baselines spanning CNN-based (U-Net \cite{ronneberger2015u}), Transformer-based (SwinUNet \cite{cao2022swin}), and SAM-adapted methods (H-SAM \cite{cheng2024unleashing}, SAN \cite{sun2024segment}, InstaSAM \cite{nam2024instasam}, MedSAM \cite{ma2024segment}). Performance is measured using Intersection over Union (IoU) and Dice coefficient (see Table~\ref{tab:pncn}).

\textbf{Zero-Shot Generalization.} To evaluate cross-domain robustness, we directly apply trained models to 12 external datasets without any fine-tuning: DSB \cite{caicedo2019nucleus}, MoNuSeg \cite{kumar2017dataset}, TNBC \cite{naylor2018segmentation}, SegPC \cite{gupta2023segpc}, GlaS \cite{sun2024segment}, Histology \cite{he2023transnuseg}, Fluorescence \cite{he2023transnuseg}, CPM15, CPM17 \cite{vu2019methods}, CryoNuSeg \cite{mahbod2021cryonuseg}, and Kumar \cite{kumar2017dataset} with two settings: Kumar-Same (test images from organ types seen during training) and Kumar-Diff (test images from entirely novel organ types). This extensive benchmark allows us to assess generalization across varying tissue types, staining protocols, imaging modalities, and organ sources. We include additional baseline methods in this evaluation to provide a more comprehensive comparison (see Section~\ref{sec: Generalization}).

\textbf{Metrics.} Following standard practice in medical image segmentation, we report Intersection over Union (IoU) and Dice Similarity Coefficient (Dice) as primary evaluation metrics.

\subsection{Implementations}
For all experiments, we employ a variant of the SAM with a ViT-B image encoder, initialized from the official checkpoint. The linear layers of model is equipped with LoRA modules \cite{hu2022lora} (default rank = 64). We use the AdamW optimizer with a base learning rate of 3e-4, a weight decay of 0.01. 
We train ours model for 10 epochs, and use an input image resolution of $224\times224$. When the original image size is slightly larger than 224 (e.g., 256), we resize it directly to $224\times224$. For images that are much larger (e.g., 512 or higher), we perform cropping to split the larger image into multiple $224\times224$ patches for both training and testing. 
The trade-off factor $\lambda_{dw}$ is 1.0. Prompts are provided in the form of positive and negative points (same as baselines). During training, each sample receives up to 3 positive and 3 negative points (same as inference phase). 
All experiments are conducted using a single V100 GPU.

\subsection{Results}

\begin{figure}
  \centering
  \includegraphics[width=1.\linewidth]{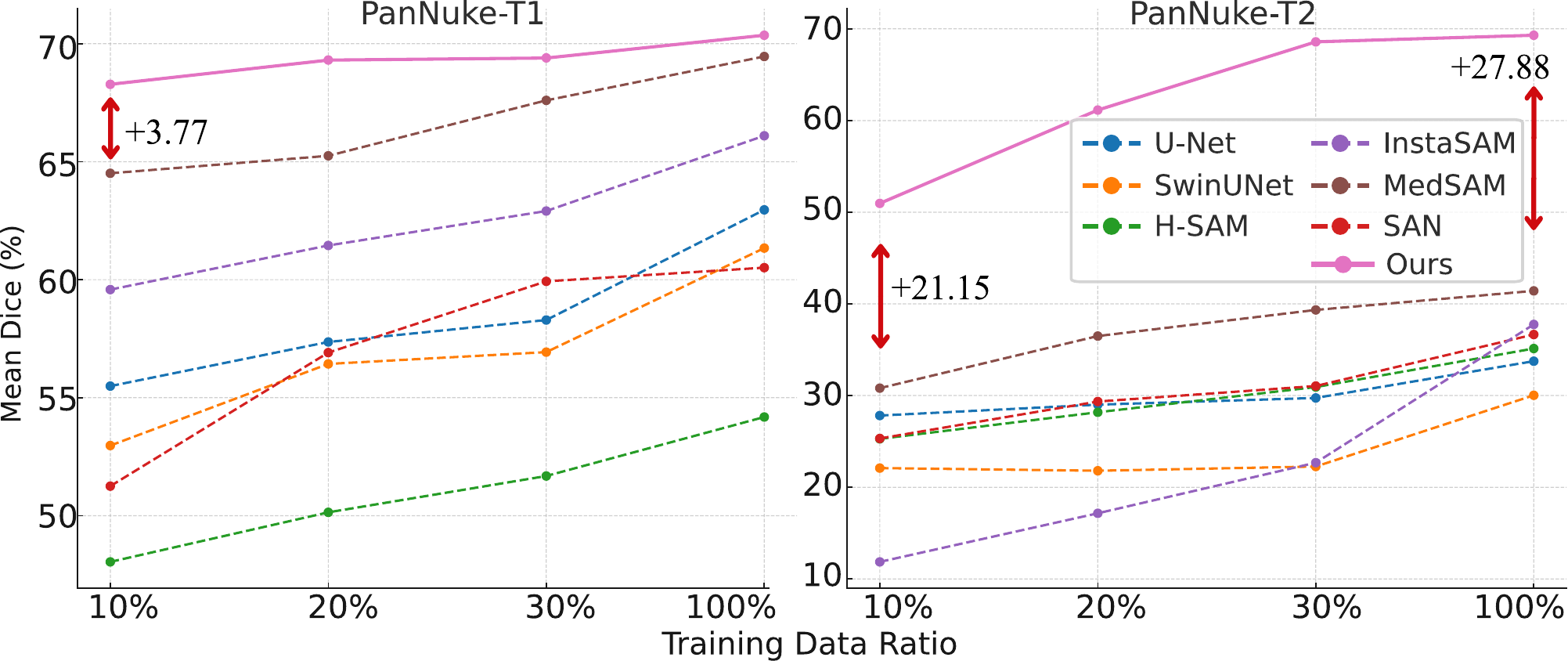}
  \caption{Performance trends of different segmentation methods on PanNuke-T1 and PanNuke-T2 under varying amounts of training data. Metric: IoU. }
  \label{fig:pn-t1-t2}
\end{figure}

\textbf{PanNuke Dataset.} Our approach is benchmarked against a suite of strong baselines, including classic architectures (U-Net \cite{ronneberger2015u}, SwinUNet \cite{cao2022swin}) and recent SAM-based methods (SAN \cite{sun2024segment}, MedSAM \cite{ma2024segment}, H-SAM \cite{cheng2024unleashing}, InstaSAM \cite{nam2024instasam}).
Quantitative results are summarized in Table~\ref{tab:pncn}, and the performance trends are visualized in Figure~\ref{fig:pn-t1-t2}. 
On the PanNuke-T1, our model achieves state-of-the-art performance across all settings. This superiority is particularly pronounced in low-data regimes, highlighting the model's robustness and data efficiency.
On the PanNuke-T2, With just 10\% of the training data, our method (50.96\% Dice) dramatically surpasses the next-best baseline, MedSAM (30.81\%).
At 100\% data, our method achieves a performance of 69.30\%, representing a 27.88\% improvement over the strongest baseline.

These results provide compelling evidence that our model excels not only in precise segmentation but also in understanding user's semantic intent. The substantial gains on the category-aware T2 task, in particular, demonstrate the model's ability to generalize from a single prompt to all instances of the same semantic class. This validates our preference learning paradigm as an effective strategy to bridge the powerful low-level perception of foundation models with high-level, user-aligned semantic understanding.

\textbf{CoNSeP Dataset.} We further extend our evaluation to the CoNSeP dataset, which entails a distinct set of challenges characterized by highly dense and overlapping nuclear structures.
Quantitative comparisons are provided in the lower half of Table~\ref{tab:pncn}.
For the binary segmentation task (CoNSeP-T1), our method consistently outperforms all baselines. Even in the 100\% data setting, where recent foundation models like MedSAM achieve competitive results (61.67\% IoU), SAMPO pushes the boundary further to 64.73\%, demonstrating superior capability in disentangling crowded instances.
The advantages of our approach are most prominent in the multi-class setting (CoNSeP-T2). 
Notably, with only 10\% of training data, our model achieves an IoU of 37.93\%, surpassing the strongest baseline (MedSAM, 21.23\% IoU) by a remarkable margin of 16.7\%. 
This performance gap suggests that in scenarios with complex tissue heterogeneity and limited annotations, purely visual-based foundation models struggle to capture semantic nuances, whereas the proposed SAMPO successfully aligns segmentation with specific semantic categories.

\subsection{Ablation Study}

\textbf{Effectiveness Analysis}. To validate the effectiveness of each core component, we conducted a comprehensive ablation study. The results, summarized in Table~\ref{tab:abla}, clearly reveal the distinct contribution of each element within our framework.

As shown in the Table~\ref{tab:abla}, removing $\mathcal{L}_{SUP}$ leads to a catastrophic performance drop across all datasets (e.g., the metric plummets to 22.34 on PanNuke-T1). This confirms its fundamental role in providing the essential guidance for pixel-level segmentation.

Building upon the supervised baseline, the preference optimization components ($\mathcal{L}_{PO1}$, $\mathcal{L}_{PO2}$) yield significant performance gains. Ablating $\mathcal{L}_{PO1}$ causes a noticeable performance decline (e.g., from 68.28 to 66.05 on PanNuke-T1), demonstrating the value of the implicit reward signals provided by its DPO objective. Similarly, the removal of $\mathcal{L}_{PO2}$ also results in a performance drop, indicating that the diverse prompt-conditioned signals it introduces are beneficial for refining the model's capabilities.

In conclusion, $\mathcal{L}_{SUP}$ provides an indispensable foundation. Upon this, $\mathcal{L}_{PO1}$ and $\mathcal{L}_{PO2}$ work synergistically to further enhance the model's final performance and robustness by bridging the "intent gap."

\begin{table*}[h]
\centering
\caption{Ablation study of key components (K1: $\mathcal{L}_{PO1}$, K2: $\mathcal{L}_{PO2}$, K3: $\mathcal{L}_{SUP}$) across PanNuke and CoNSeP. Metric: IoU and Dice Score. }
\label{tab:abla}
\begin{tabular}{ccc|cc|cc|cc|cc}
\hline
\multirow{2}{*}{K1}       & \multirow{2}{*}{K2} & \multirow{2}{*}{K3} & \multicolumn{2}{c|}{PanNuke-T1} & \multicolumn{2}{c|}{PanNuke-T2} & \multicolumn{2}{c|}{CoNSeP-T1} & \multicolumn{2}{c}{CoNSeP-T2} \\ \cline{4-11} 
                          & & & \multicolumn{1}{c}{IoU} & \multicolumn{1}{c|}{Dice} & \multicolumn{1}{c}{IoU} & \multicolumn{1}{c|}{Dice} & IoU         & Dice   & IoU               & Dice       \\ \hline
\checkmark & \checkmark & \checkmark     & \textbf{68.28}    & \textbf{81.15}    & \textbf{50.96}    & \textbf{67.51}    & \textbf{64.73}    & \textbf{78.17}    & \textbf{46.69}    & \textbf{61.56}   \\ \hline
\ding{55}    & \checkmark & \checkmark   & 66.05             & 79.55             & 47.91             & 64.78             & 61.35             & 76.05             & 41.83             & 58.99             \\
\checkmark & \ding{55}    & \checkmark   & 65.42             & 79.10             & 48.53             & 65.35             & 60.21             & 75.16             & 43.57             & 60.70             \\
\checkmark & \checkmark & \ding{55}      & 22.34             & 36.52             & 34.51             & 51.31             & 32.49             & 49.05             & 25.62             & 40.79             \\ \hline
\ding{55}    & \ding{55}    & \checkmark & 65.29             & 79.00             & 46.99             & 63.94             & 59.14             & 74.32             & 38.41             & 55.50             \\
\checkmark & \ding{55}    & \ding{55}    & 25.29             & 40.37             & 34.72             & 51.54             & 37.73             & 54.79             & 24.75             & 39.68             \\ \hline
\end{tabular}\end{table*}

\textbf{Impact of $\lambda_{\mathrm{dw}}$}. 
We investigate the impact of the preference weight $\lambda_{dw}$ in Table~\ref{tab:dw}. An overly low $\lambda_{dw}$ provides a weak preference signal, whereas an excessively high value leads to overfitting on preference pairs, thereby compromising pixel-level accuracy. Empirically, we find that $\lambda_{dw} = 1.0$ yields optimal outcomes on all benchmarks. This result underscores the importance of carefully balancing this weight to align the model with user preferences without degrading its core segmentation capabilities.

\begin{table*}[h]
\centering
\caption{The effect of varying the preference optimization weight $\lambda_{dw}$. Metric: IoU and Dice Score. }
\label{tab:dw}
\begin{tabular}{c|cl|cl|cc|cc}
\hline
\multirow{2}{*}{$\lambda_{dw}$} & \multicolumn{2}{c|}{PanNuke-T1}  & \multicolumn{2}{c|}{PanNuke-T2}  & \multicolumn{2}{c|}{CoNSeP-T1}  & \multicolumn{2}{c}{CoNSeP-T2}  \\ \cline{2-9} 
                                & \multicolumn{1}{c}{IoU}          & Dice  & \multicolumn{1}{c}{IoU}  & Dice  & \multicolumn{1}{c}{IoU} & \multicolumn{1}{c|}{Dice} & \multicolumn{1}{c}{IoU} & \multicolumn{1}{c}{Dice} \\ \hline
0.10                            & 67.97          & 80.93          & 50.64          & 67.23          & 63.74              & 77.86              & 43.83              & 60.95                    \\
0.50                            & 67.50          & 80.60          & 50.85          & 67.42          & 64.26              & \textbf{78.24}     & 44.52              & 61.61           \\
1.00                            & \textbf{68.28} & \textbf{81.15} & \textbf{50.96} & \textbf{67.51} & \textbf{64.73}     & 78.17              & \textbf{46.69}     & \textbf{61.56}                    \\
2.00                            & 66.96          & 80.21          & 46.18          & 63.18          & 63.45              & 77.64              & 45.36              & 62.41                    \\ \hline
\end{tabular}
\end{table*}

\textbf{LoRA Rank Selection}. 
We integrate LoRA into the SAM for parameter-efficient fine-tuning and evaluate the impact of its rank in Table~\ref{tab:rank}. The performance scales with the rank, especially on complex datasets like PanNuke-T2 and Synapse. While lower ranks such as 4 or 8 cause significant performance drops due to insufficient model capacity, a rank of 64 consistently delivers the best results across all benchmarks without showing signs of overfitting. This confirms that a rank of 64 offers a favorable trade-off between model accuracy and parameter efficiency.

\begin{table*}[h]
\centering
\caption{Ablation about LoRA Rank. All results are obtained by training on 10\% of the training data. Metric: IoU and Dice Score. }
\label{tab:rank}
\begin{tabular}{c|cl|cl|cl|cl}
\hline
\multirow{2}{*}{Rank} & \multicolumn{2}{c|}{PanNuke-T1} & \multicolumn{2}{c|}{PanNuke-T2} & \multicolumn{2}{c|}{CoNSeP-T1} & \multicolumn{2}{c}{CoNSeP-T2}  \\ \cline{2-9} 
                      & \multicolumn{1}{c}{IoU}  & Dice               & \multicolumn{1}{c}{IoU}  & Dice              & \multicolumn{1}{c}{IoU} & Dice             & \multicolumn{1}{c}{IoU}   & Dice             \\ \hline
4                     & 65.71                    & 79.31              & 46.97                    & 63.92             & 62.35                   & 76.81            & 42.06                     & 59.21            \\
8                     & 66.17                    & 79.64              & 47.63                    & 64.53             & 62.78                   & 77.13            & 43.19                     & 60.33            \\
32                    & 67.61                    & 80.68              & 50.36                    & 66.99             & 63.40                   & 77.60            & 44.52                     & 61.61            \\
64                    & \textbf{68.28}           & \textbf{81.15}     & \textbf{50.96}           & \textbf{67.51}    & \textbf{64.73}          & \textbf{78.17}   & \textbf{46.69}            & \textbf{61.56}   \\ \hline
\end{tabular}
\end{table*}

\textbf{Point Prompt Sensitivity Analysis}. 
We conducted a point prompt sensitivity analysis to systematically compare the performance of our method (SAMPO) against SAM2 across various point prompt configurations. The analysis was performed on the PanNuke-T1 and T2 tasks by varying the number of positive and negative points.

\begin{figure}[h]
    \centering
    \includegraphics[width=1\linewidth]{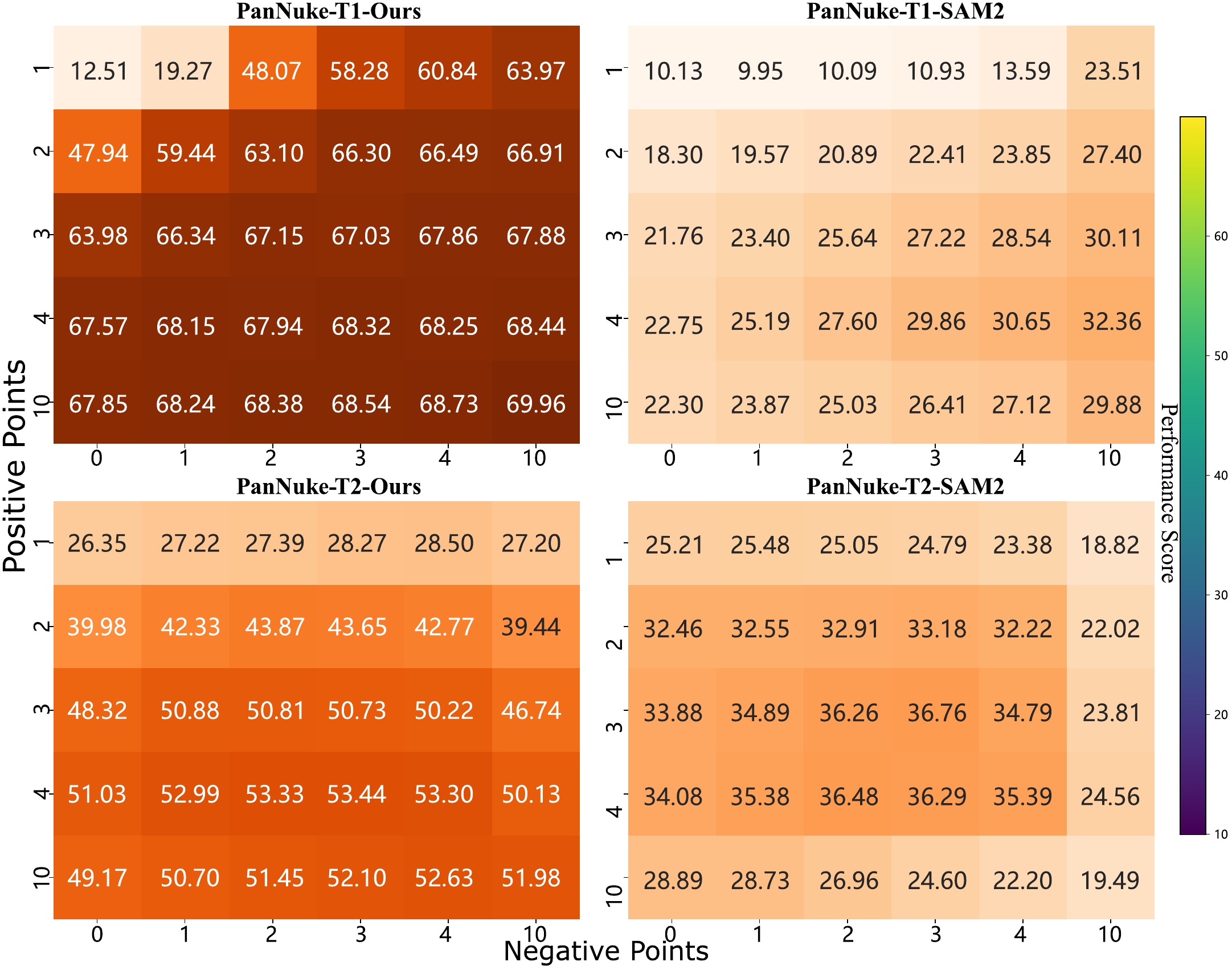}
    \caption{Point prompt sensitivity analysis on the PanNuke T1 and T2. Heatmaps show the Dice scores of SAMPO and the original SAM2 across varying numbers of positive (1–10) and negative (0–10) point prompts. A unified color scale is used for direct comparison.}
    \label{fig:point}
\end{figure}

As shown in Figure~\ref{fig:point}, the results demonstrate that SAMPO significantly outperforms SAM2 across nearly all prompt combinations. 
More importantly, our method is less sensitive to variations in the number of prompt points, demonstrating greater robustness and stability. This indicates that our preference optimization framework effectively enhances the model's understanding of user intent, making it more reliable for practical applications.

\subsection{Qualitative Results}

\begin{figure*}[h]
    \centering
    \includegraphics[width=1\linewidth]{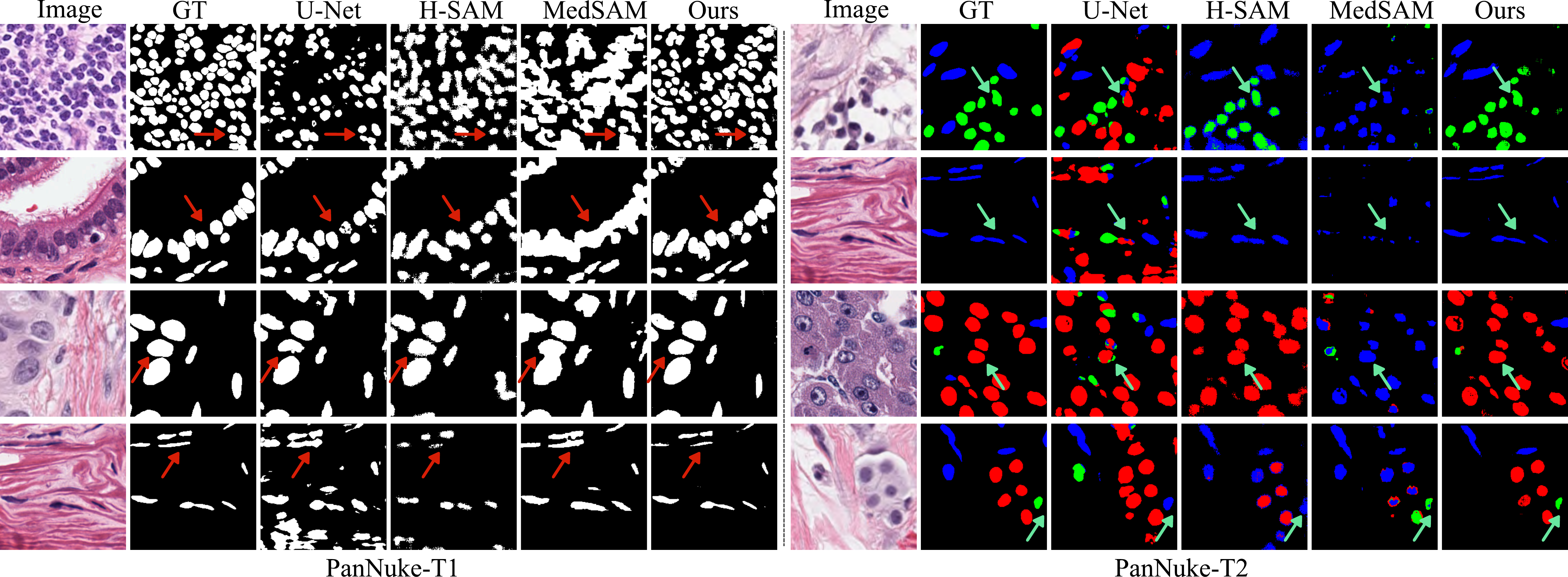}
    \caption{The qualitative results in the PanNuke dataset. The arrow indicates the key differences. }
    \label{fig:vis-pn-seg}
\end{figure*}

\begin{figure*}[h]
    \centering
    \includegraphics[width=1\linewidth]{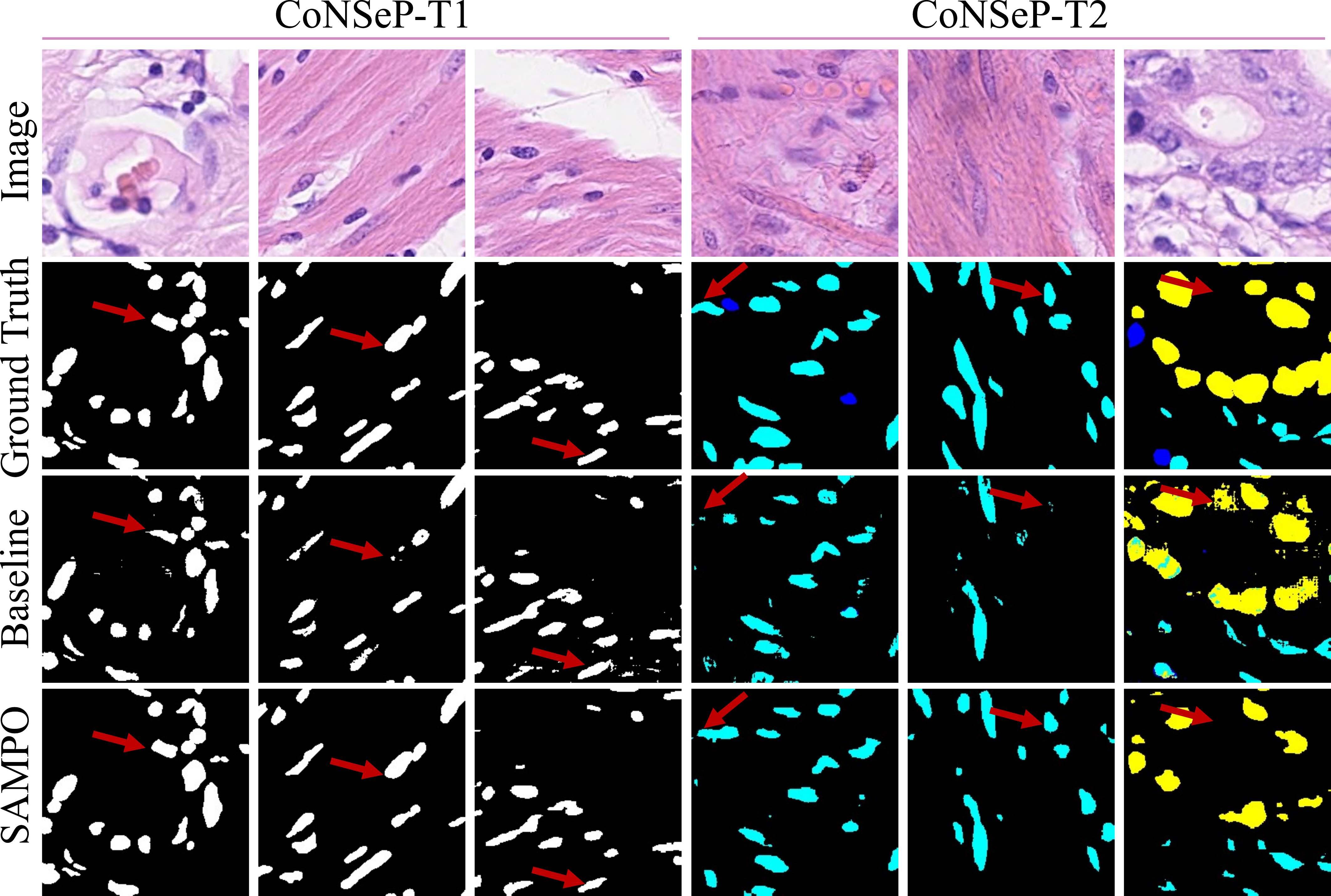}
    \caption{The qualitative results in the CoNSeP dataset. The arrow indicates the key differences. Baseline: MedSAM. }
    \label{fig:vis-cn-seg}
\end{figure*}

We conduct experiments using 10\% of the training data on two datasets for qualitative analysis. As shown in Figure~\ref{fig:vis-pn-seg}, compared to U-Net, H-SAM, and MedSAM, SAMPO required only a few point prompts (3 positive and 3 negative) to achieve the most precise boundary delineation and morphological fidelity in regions with densely packed nuclei. More importantly, SAMPO can automatically generalize prompt information from a single target to all same-type cells within the field of view, thereby eliminating the tedious task of instance-by-instance prompting. Additionally, results on another dataset, CoNSeP, are provided as shown in Figure~\ref{fig:vis-cn-seg}, further demonstrating the robustness and generalization ability of SAMPO across different tissue types.

\subsection{Generalization Analysis}
\label{sec: Generalization}

\begin{figure*}
    \centering
    \includegraphics[width=0.8\textwidth]{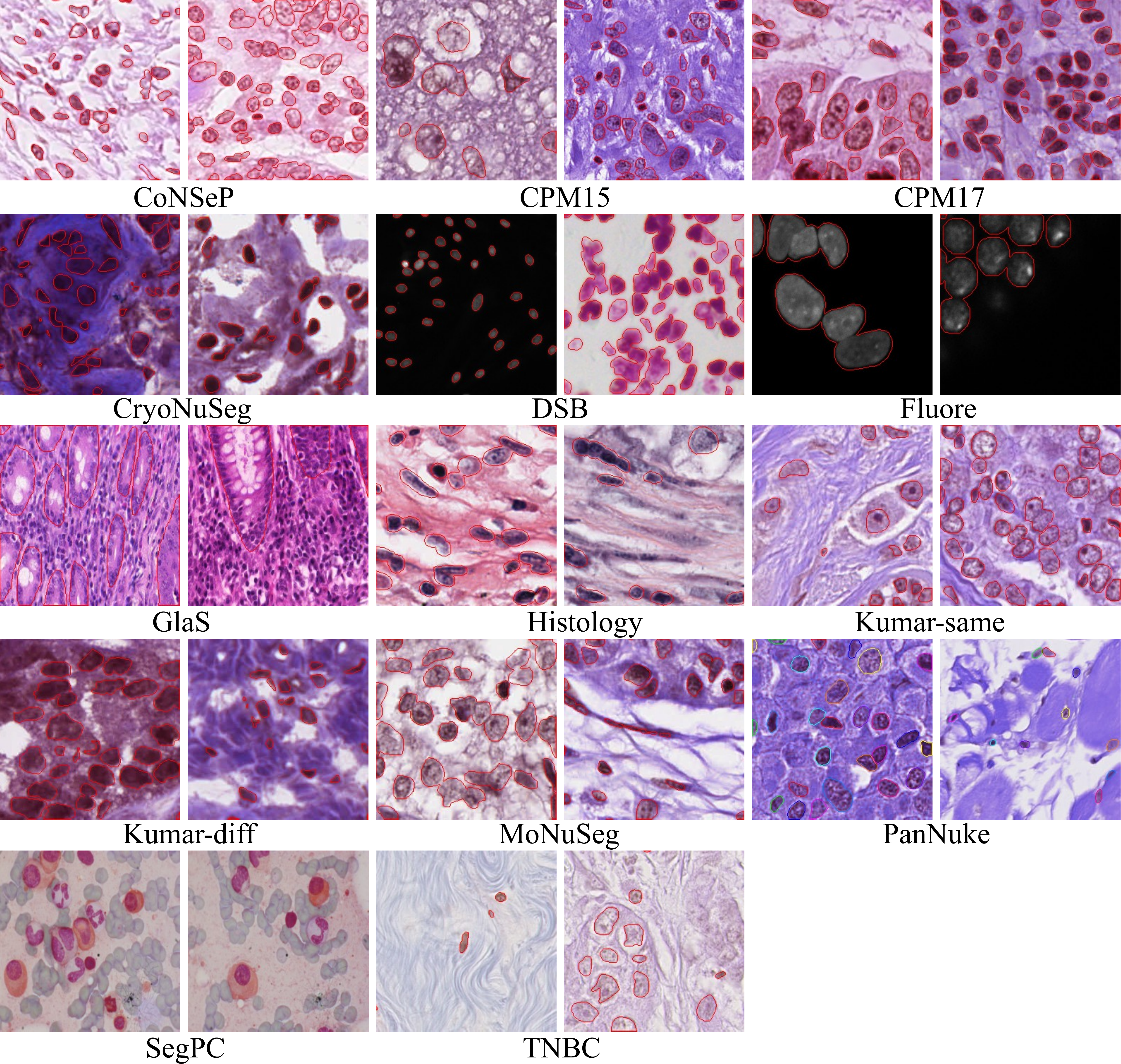}
    \caption{Sample cropped regions extracted from each of the five nuclear instance segmentation datasets used in our experiments.}
    \label{fig:data-overview}
\end{figure*}

To rigorously evaluate the generalization capability of SAMPO, we conduct extensive experiments across multiple dimensions: nucleus density variation, cross-category performance, 
and cross-tissue adaptation. We compare SAMPO against a comprehensive set of baseline methods spanning classical encoder-decoder architectures 
(U-Net \cite{ronneberger2015u}, UNet++ \cite{zhou2022domain}, Attention-UNet \cite{schlemper2019attention}, TransUNet \cite{CHEN2024103280}, nnU-Net \cite{isensee2020nnunet}), 
SAM-based adaptation methods (SAMed \cite{zhang2023customized}, SAM \cite{SAM}, Med-SA \cite{WU2025103547}, All-in-SAM \cite{cui2024all}, H-SAM \cite{cheng2024unleashing}), 
and recent promptable segmentation approaches (CellSAM \cite{israel2025cellsam}, SAC \cite{na2025segment}, PromptNucSeg \cite{shui2024unleashing}, SAM2 \cite{ravi2024sam2}, 
SAM3 \cite{carion2025sam3segmentconcepts}). This extensive comparison enables us to assess not only the absolute performance but also the robustness of 
our approach across diverse evaluation dimensions. An overview of the external datasets used for zero-shot evaluation is provided in Figure~\ref{fig:data-overview}.

\subsubsection{Impact of Nucleus Density}
\label{sec:density}

\begin{figure*}
    \centering
    \includegraphics[width=1\linewidth]{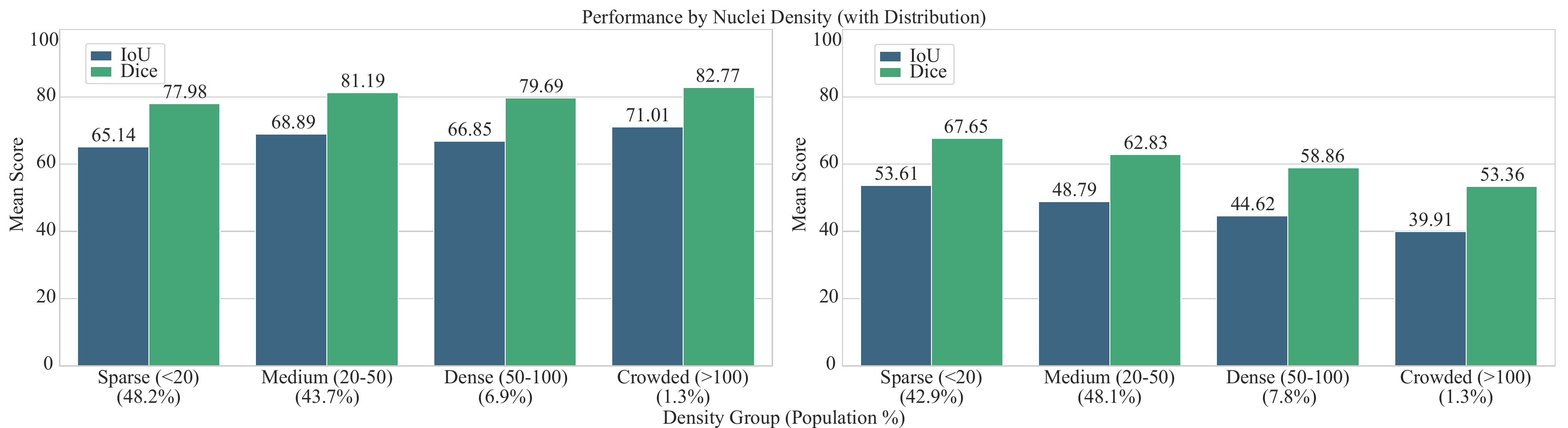}
    \caption{Impact of nucleus density on segmentation performance. Results are reported on PanNuke for both Universal Nuclei Segmentation (T1) and Category-Specific Nuclei Segmentation (T2) tasks. Percentage values in parentheses indicate the proportion of test images in each density category.}
    \label{fig:density}
\end{figure*}

Histopathological images exhibit substantial variation in nucleus density, ranging from sparsely populated stromal regions to densely packed tumor nests. To investigate how this variation affects segmentation performance, we stratify the test images into four density categories: Sparse ($<$20 nuclei), Medium (20--50 nuclei), Dense (50--100 nuclei), and Crowded ($>$100 nuclei). We evaluate SAMPO under both task configurations on the PanNuke dataset.

Figure~\ref{fig:density} presents the results across density levels. For the Universal Nuclei Segmentation task (T1), we observe a notable trend: performance improves with increasing nucleus density, with IoU rising from 65.14\% (Sparse) to 71.01\% (Crowded). This counter-intuitive finding can be attributed to the nature of the prompting mechanism in T1. With more nuclei present in the image, the sparse point prompts provide a more representative sampling of the "nucleus-ness" pattern, enabling the model to better capture the underlying structural distribution. Additionally, denser regions often exhibit more homogeneous morphological characteristics, reducing ambiguity in pattern inference.

In contrast, the Category-Specific Nuclei Segmentation task (T2) exhibits the opposite trend: performance decreases with increasing density, with IoU dropping from 53.61\% (Sparse) to 39.91\% (Crowded). This degradation reflects the inherent difficulty of semantic disambiguation in crowded scenes. When multiple nucleus categories are densely interleaved, distinguishing the target category from visually similar neighbors becomes increasingly challenging. The model must simultaneously reason about morphological features and spatial context to resolve category boundaries, a task that becomes more ambiguous as nuclei of different types cluster together.

These findings highlight a fundamental distinction between the two task formulations: T1 benefits from statistical redundancy in dense images, while T2 is constrained by the increased complexity of semantic reasoning in crowded environments. This analysis suggests that future improvements in T2 performance may require enhanced mechanisms for capturing fine-grained categorical distinctions in high-density regions.

\subsubsection{Cross-Category Performance}
The Category-Specific Nuclei Segmentation task (T2) requires the model to segment nuclei belonging to a particular semantic category based on sparse exemplar prompts. This poses a significant challenge, as nucleus categories can exhibit substantial intra-class morphological variability while maintaining subtle inter-class differences. To provide a fine-grained understanding of model behavior, we analyze the per-category performance breakdown on both PanNuke and CoNSeP test sets (Table~\ref{tab:nucleus}).

On PanNuke, SAMPO demonstrates consistent superiority across all five nucleus categories. Most notably, we observe substantial improvements in categories that baseline methods struggle with. For Inflammatory Nuclei, SAMPO achieves 56.11\% IoU compared to 26.61\% for MedSAM and 13.53\% for H-SAM, representing an absolute improvement of over 29 pp. Similarly, for Dead Nuclei, a category where U-Net and H-SAM fail entirely, SAMPO achieves 42.61\% IoU, demonstrating robust handling of challenging morphological patterns. For Neoplastic and Epithelial Nuclei, which are typically more prominent and morphologically distinctive, SAMPO maintains strong performance with IoU scores of 53.50\% and 53.39\%, respectively.

The CoNSeP dataset presents additional challenges with its finer-grained category distinctions in colorectal adenocarcinoma tissue. Again, SAMPO consistently outperforms all baselines across categories. The improvement is particularly pronounced for Miscellaneous Nuclei (43.87\% vs. 26.55\% for MedSAM) and Inflammatory Nuclei (46.35\% vs. 29.39\% for MedSAM). For Epithelial and Spindle-shaped Nuclei, while the margins are smaller, SAMPO still maintains the leading performance.

These results validate SAMPO's ability to learn transferable categorical representations from sparse prompts. The consistent improvements across diverse nucleus types, ranging from the morphologically distinct (Neoplastic, Epithelial) to the challenging and underrepresented (Dead, Miscellaneous), indicate that our prompt-aware learning framework effectively captures both common structural patterns and category-specific discriminative features.

\begin{table*}[!ht]
\centering
\caption{Per-category performance analysis on PanNuke and CoNSeP. Best results for each tissue type are shown in \textbf{bold}.}
\label{tab:nucleus}
\small
\begin{tabular}{c|cccccccc}
\hline
\multirow{3}{*}{Class} & \multicolumn{8}{c}{PanNuke-T2} \\ \cline{2-9} 
 & \multicolumn{2}{c|}{U-Net}                           & \multicolumn{2}{c|}{H-SAM}                           & \multicolumn{2}{c|}{MedSAM}                          & \multicolumn{2}{c}{Ours}                            \\ \cline{2-9} 
 & \multicolumn{1}{c}{IoU} & \multicolumn{1}{c|}{Dice}  & \multicolumn{1}{c}{IoU} & \multicolumn{1}{c|}{Dice}  & \multicolumn{1}{c}{IoU} & \multicolumn{1}{c|}{Dice}  & \multicolumn{1}{c}{IoU} & \multicolumn{1}{c}{Dice}  \\ \hline
Neoplastic Nuclei       & 45.38  & \multicolumn{1}{c|}{59.97} & 42.80  & \multicolumn{1}{c|}{55.55} & 36.37  & \multicolumn{1}{c|}{50.93} & \textbf{53.50}   & \multicolumn{1}{c}{\textbf{68.05}} \\
Inflammatory Nuclei     & 25.71  & \multicolumn{1}{c|}{37.08} & 13.53  & \multicolumn{1}{c|}{19.09} & 26.61  & \multicolumn{1}{c|}{38.90} & \textbf{56.11}   & \multicolumn{1}{c}{\textbf{68.93}} \\
Connective Nuclei       & 19.03  & \multicolumn{1}{c|}{29.72} & 14.80  & \multicolumn{1}{c|}{21.11} & 29.11  & \multicolumn{1}{c|}{41.90} & \textbf{43.99}   & \multicolumn{1}{c}{\textbf{58.40}}  \\
Dead Nuclei             & -      & \multicolumn{1}{c|}{-}     & -      & \multicolumn{1}{c|}{-}     & 17.51  & \multicolumn{1}{c|}{27.02} & \textbf{42.61}   & \multicolumn{1}{c}{\textbf{56.55}} \\
Epithelial Nuclei       & -      & \multicolumn{1}{c|}{-}     & 40.29  & \multicolumn{1}{c|}{52.36} & 35.16  & \multicolumn{1}{c|}{49.71} & \textbf{53.39}   & \multicolumn{1}{c}{\textbf{67.88}} \\ \hline

\multirow{3}{*}{Class} & \multicolumn{8}{c}{CoNSeP-T2} \\ \cline{2-9} 
 & \multicolumn{2}{c|}{U-Net} & \multicolumn{2}{c|}{SwinUNet} & \multicolumn{2}{c|}{MedSAM} & \multicolumn{2}{c}{Ours} \\ \cline{2-9} 
 & \multicolumn{1}{c}{IoU}    & \multicolumn{1}{c|}{Dice}     & \multicolumn{1}{c}{IoU}     & \multicolumn{1}{c|}{Dice}  & \multicolumn{1}{c}{IoU} & \multicolumn{1}{c|}{Dice}  & \multicolumn{1}{c}{IoU} & \multicolumn{1}{c}{Dice}   \\ \hline
Miscellaneous Nuclei   &  1.08   & \multicolumn{1}{c|}{ 2.09}  & -     & \multicolumn{1}{c|}{-}     & 26.55   & \multicolumn{1}{c|}{35.90}    & \textbf{43.87}   & \textbf{57.56} \\
Inflammatory Nuclei    &  9.80   & \multicolumn{1}{c|}{17.42}  &  5.86 & \multicolumn{1}{c|}{10.34} & 29.39   & \multicolumn{1}{c|}{41.83}    & \textbf{46.35}   & \textbf{61.39} \\
Epithelial Nuclei      & 37.67   & \multicolumn{1}{c|}{52.31}  & 39.20 & \multicolumn{1}{c|}{54.72} & 41.81   & \multicolumn{1}{c|}{56.36}    & \textbf{44.32}   & \textbf{59.73} \\
Spindle-shaped Nuclei  &  3.52   & \multicolumn{1}{c|}{ 6.69}  & 24.19 & \multicolumn{1}{c|}{36.84} & 30.05   & \multicolumn{1}{c|}{42.45}    & \textbf{30.68}   & \textbf{45.05} \\ \hline
\end{tabular}
\end{table*}

\subsubsection{Cross-Tissue Performance}

Histopathological images from different tissue types exhibit distinct morphological characteristics, including variations in cellular density, nuclear shape, staining intensity, and tissue architecture. To evaluate whether SAMPO maintains consistent performance across this biological diversity, we analyze the per-tissue performance breakdown on the PanNuke test set, which encompasses 19 different tissue types from multiple organ systems.

Table~\ref{tab:tissue} presents the comprehensive results for both T1 and T2 tasks across all 19 tissue types. For the Universal Nuclei Segmentation task (T1), SAMPO achieves the best performance on 17 out of 19 tissue types, with an average IoU of 67.56\% compared to 63.50\% for MedSAM and 54.62\% for U-Net. The improvement is particularly notable in challenging tissue types such as Head\&Neck (60.90\% vs. 49.57\% for MedSAM), Kidney (68.35\% vs. 61.94\%), and Skin (66.25\% vs. 59.07\%). These tissues often contain heterogeneous cellular compositions and complex architectural patterns, where SAMPO's prompt-aware mechanism proves especially beneficial. Notably, MedSAM achieves slightly better performance on Cervix (66.62\% vs. 65.79\%) and Testis (65.72\% vs. 64.44\%), suggesting that certain tissue-specific patterns may benefit from different adaptation strategies.

For the Category-Specific Nuclei Segmentation task (T2), the performance gap between SAMPO and baseline methods becomes even more pronounced. SAMPO achieves an average IoU of 50.67\%, substantially outperforming MedSAM (28.47\%), H-SAM (22.02\%), and U-Net (20.31\%). This represents an absolute improvement of over 22 percentage points compared to the best baseline. The improvement is consistent across all 19 tissue types, with particularly large gains observed in Uterus (39.91\% vs. 14.85\% for H-SAM), Adrenal (56.04\% vs. 31.57\% for MedSAM), and Lung (45.05\% vs. 20.62\% for MedSAM). The uniformly strong performance across diverse tissue types demonstrates that SAMPO's prompt refinement mechanism successfully captures tissue-agnostic features for category-specific segmentation.

Interestingly, we observe that certain tissue types consistently pose greater challenges across all methods. For instance, Uterus shows relatively lower T2 performance even for SAMPO (39.91\% IoU), and Testis exhibits the smallest improvement margin in T1. These observations suggest potential directions for future work in developing tissue-adaptive mechanisms.

\begin{table*}[!ht]
\centering
\setlength{\tabcolsep}{3pt} 
\caption{Per-tissue performance analysis on PanNuke. Results are reported for both Universal Nuclei Segmentation (T1) and Category-Specific Nuclei Segmentation (T2) tasks across all 19 tissue types. Best results for each tissue type are shown in \textbf{bold}.}
\label{tab:tissue}
\begin{tabular}{c|cccccccc|cccccccc}
\hline
\multirow{3}{*}{Tissue} & \multicolumn{8}{c|}{PanNuke-T1} & \multicolumn{8}{c}{PanNuke-T2} \\ \cline{2-17} 
                        & \multicolumn{2}{c|}{U-Net} & \multicolumn{2}{c|}{InstaSAM} & \multicolumn{2}{c|}{MedSAM} & \multicolumn{2}{c|}{SAMPO} & \multicolumn{2}{c|}{U-Net} & \multicolumn{2}{c|}{H-SAM} & \multicolumn{2}{c|}{MedSAM} & \multicolumn{2}{c}{SAMPO} \\ \cline{2-17} 
                        & IoU & \multicolumn{1}{c|}{Dice}  & IoU & \multicolumn{1}{c|}{Dice}  & IoU & \multicolumn{1}{c|}{Dice}  & IoU & Dice & IoU & \multicolumn{1}{c|}{Dice}  & IoU & \multicolumn{1}{c|}{Dice}  & IoU & \multicolumn{1}{c|}{Dice}  & IoU & Dice \\ \hline
Adrenal                 & 57.22 & \multicolumn{1}{c|}{70.43} & 42.57 & \multicolumn{1}{c|}{58.01} & 65.49 & \multicolumn{1}{c|}{78.03} & \textbf{67.87} & \textbf{80.06} & 17.63 & \multicolumn{1}{c|}{24.87} & 23.89 & \multicolumn{1}{c|}{31.07} & 31.57 & \multicolumn{1}{c|}{44.58} & \textbf{56.04} & \textbf{69.95} \\
Bile Duct               & 59.86 & \multicolumn{1}{c|}{73.15} & 46.25  & \multicolumn{1}{c|}{62.24} & 64.92 & \multicolumn{1}{c|}{78.02} & \textbf{70.27} & \textbf{82.06}  & 24.01 & \multicolumn{1}{c|}{34.01} & 21.22 & \multicolumn{1}{c|}{29.84} & 33.85 & \multicolumn{1}{c|}{48.19} & \textbf{46.96} & \textbf{61.58} \\
Bladder                 & 50.27 & \multicolumn{1}{c|}{64.13} & 42.13  & \multicolumn{1}{c|}{58.46} & 59.82 & \multicolumn{1}{c|}{73.74} & \textbf{67.97} & \textbf{80.46}  & 23.95 & \multicolumn{1}{c|}{33.47} & 24.46 & \multicolumn{1}{c|}{32.35} & 28.08 & \multicolumn{1}{c|}{40.00}    & \textbf{50.04} & \textbf{64.81} \\
Breast                  & 59.62 & \multicolumn{1}{c|}{73.46} & 49.99  & \multicolumn{1}{c|}{66.11} & 66.77 & \multicolumn{1}{c|}{79.58} & \textbf{68.50}  & \textbf{80.84}  & 19.69 & \multicolumn{1}{c|}{27.16} & 22.19 & \multicolumn{1}{c|}{29.25} & 30.11 & \multicolumn{1}{c|}{43.28} & \textbf{55.30}  & \textbf{69.24} \\
Cervix                  & 50.6  & \multicolumn{1}{c|}{63.26} & 45.31  & \multicolumn{1}{c|}{60.03} & \textbf{66.62} & \multicolumn{1}{c|}{\textbf{78.50}}  & 65.79 & 78.15  & 22.64 & \multicolumn{1}{c|}{31.76} & 23.20  & \multicolumn{1}{c|}{30.86} & 27.75 & \multicolumn{1}{c|}{40.48} & \textbf{50.27} & \textbf{64.59} \\
Colon                   & 49.82 & \multicolumn{1}{c|}{64.32} & 44.43  & \multicolumn{1}{c|}{60.19} & 60.54 & \multicolumn{1}{c|}{74.36} & \textbf{62.06} & \textbf{75.91}  & 15.29 & \multicolumn{1}{c|}{22.24} & 22.91 & \multicolumn{1}{c|}{30.77} & 29.36 & \multicolumn{1}{c|}{42.43} & \textbf{50.73} & \textbf{65.36} \\
Esophagus               & 54.55 & \multicolumn{1}{c|}{68.63} & 46.51  & \multicolumn{1}{c|}{62.24} & 65.16 & \multicolumn{1}{c|}{77.89} & \textbf{67.68} & \textbf{79.79}  & 19.04 & \multicolumn{1}{c|}{26.23} & 24.69 & \multicolumn{1}{c|}{32.94} & 33.22 & \multicolumn{1}{c|}{46.63} & \textbf{53.04} & \textbf{66.74} \\
Head\&Neck              & 41.01 & \multicolumn{1}{c|}{54.80}  & 33.90   & \multicolumn{1}{c|}{48.79} & 49.57 & \multicolumn{1}{c|}{64.41} & \textbf{60.90}  & \textbf{75.11}  & 16.20  & \multicolumn{1}{c|}{23.35} & 19.43 & \multicolumn{1}{c|}{26.75} & 26.05 & \multicolumn{1}{c|}{39.12} & \textbf{49.89} & \textbf{63.91} \\
Kidney                  & 42.97 & \multicolumn{1}{c|}{55.84} & 37.46  & \multicolumn{1}{c|}{51.89} & 61.94 & \multicolumn{1}{c|}{76.10}  & \textbf{68.35} & \textbf{80.51}  & 24.36 & \multicolumn{1}{c|}{34.25} & 22.51 & \multicolumn{1}{c|}{28.40}  & 29.85 & \multicolumn{1}{c|}{42.57} & \textbf{52.01} & \textbf{66.08} \\
Liver                   & 64.54 & \multicolumn{1}{c|}{77.40}  & 47.13  & \multicolumn{1}{c|}{63.29} & 70.99 & \multicolumn{1}{c|}{82.67} & \textbf{72.23} & \textbf{83.33}  & 19.51 & \multicolumn{1}{c|}{26.6}  & 17.92 & \multicolumn{1}{c|}{23.97} & 28.04 & \multicolumn{1}{c|}{39.20}  & \textbf{46.49} & \textbf{60.53} \\
Lung                    & 55.14 & \multicolumn{1}{c|}{69.33} & 42.49  & \multicolumn{1}{c|}{58.16} & 60.57 & \multicolumn{1}{c|}{74.06} & \textbf{64.88} & \textbf{78.24}  & 21.58 & \multicolumn{1}{c|}{31.20}  & 11.90  & \multicolumn{1}{c|}{16.24} & 20.62 & \multicolumn{1}{c|}{31.76} & \textbf{45.05} & \textbf{58.67} \\
Ovarian                 & 56.39 & \multicolumn{1}{c|}{70.89} & 48.97  & \multicolumn{1}{c|}{65.08} & 63.53 & \multicolumn{1}{c|}{77.08} & \textbf{67.59} & \textbf{80.16}  & 25.28 & \multicolumn{1}{c|}{34.99} & 25.60  & \multicolumn{1}{c|}{33.88} & 30.43 & \multicolumn{1}{c|}{43.60}  & \textbf{55.04} & \textbf{69.19} \\
Pancreatic              & 54.78 & \multicolumn{1}{c|}{69.28} & 44.01  & \multicolumn{1}{c|}{59.91} & 60.65 & \multicolumn{1}{c|}{74.34} & \textbf{69.92} & \textbf{81.87}  & 19.37 & \multicolumn{1}{c|}{28.59} & 23.70  & \multicolumn{1}{c|}{31.92} & 33.03 & \multicolumn{1}{c|}{47.14} & \textbf{54.03} & \textbf{68.85} \\
Prostate                & 55.97 & \multicolumn{1}{c|}{69.98} & 47.80   & \multicolumn{1}{c|}{63.86} & 65.41 & \multicolumn{1}{c|}{78.30}  & \textbf{67.99} & \textbf{80.52}  & 23.10  & \multicolumn{1}{c|}{33.18} & 21.04 & \multicolumn{1}{c|}{28.73} & 35.76 & \multicolumn{1}{c|}{50.67} & \textbf{61.70}  & \textbf{73.54} \\
Skin                    & 45.48 & \multicolumn{1}{c|}{59.57} & 43.20   & \multicolumn{1}{c|}{57.38} & 59.07 & \multicolumn{1}{c|}{72.48} & \textbf{66.25} & \textbf{78.19}  & 15.30  & \multicolumn{1}{c|}{22.71} & 33.38 & \multicolumn{1}{c|}{42.45} & 23.34 & \multicolumn{1}{c|}{34.86} & \textbf{52.71} & \textbf{66.39} \\
Stomach                 & 66.29 & \multicolumn{1}{c|}{79.43} & 53.01  & \multicolumn{1}{c|}{69.20}  & 70.65 & \multicolumn{1}{c|}{82.56} & \textbf{74.34} & \textbf{85.15}  & 23.96 & \multicolumn{1}{c|}{34.36} & 16.14 & \multicolumn{1}{c|}{22.57} & 21.66 & \multicolumn{1}{c|}{32.39} & \textbf{44.29} & \textbf{57.84} \\
Testis                  & 57.75 & \multicolumn{1}{c|}{71.71} & 48.33  & \multicolumn{1}{c|}{64.50}  & \textbf{65.72} & \multicolumn{1}{c|}{\textbf{78.91}} & 64.44 & 77.68  & 22.03 & \multicolumn{1}{c|}{31.59} & 23.81 & \multicolumn{1}{c|}{32.28} & 32.97 & \multicolumn{1}{c|}{46.99} & \textbf{43.98} & \textbf{58.41} \\
Thyroid                 & 58.34 & \multicolumn{1}{c|}{70.38} & 42.34  & \multicolumn{1}{c|}{57.89} & 66.73 & \multicolumn{1}{c|}{79.49} & \textbf{70.26} & \textbf{82.12}  & 19.96 & \multicolumn{1}{c|}{28.53} & 25.52 & \multicolumn{1}{c|}{33.77} & 32.82 & \multicolumn{1}{c|}{46.15} & \textbf{55.21} & \textbf{68.79} \\
Uterus                  & 57.26 & \multicolumn{1}{c|}{72.03} & 49.41  & \multicolumn{1}{c|}{65.68} & 62.35 & \multicolumn{1}{c|}{76.2}  & \textbf{66.35} & \textbf{79.40}   & 12.94 & \multicolumn{1}{c|}{19.15} & 14.85 & \multicolumn{1}{c|}{19.36} & 12.35 & \multicolumn{1}{c|}{19.48} & \textbf{39.91} & \textbf{52.32} \\ \hline

Average                 & \multicolumn{1}{c}{54.62} & \multicolumn{1}{c|}{68.32} & \multicolumn{1}{c}{55.32} & \multicolumn{1}{c|}{70.43} & \multicolumn{1}{c}{63.50} & \multicolumn{1}{c|}{76.67} & \multicolumn{1}{c}{\textbf{67.56}} & \multicolumn{1}{c|}{\textbf{79.98}} & \multicolumn{1}{c}{20.31} & \multicolumn{1}{c|}{28.85} & \multicolumn{1}{c}{22.02} & \multicolumn{1}{c|}{29.34} & \multicolumn{1}{c}{28.47} & \multicolumn{1}{c|}{41.03} & \multicolumn{1}{c}{\textbf{50.67}} & \multicolumn{1}{c}{\textbf{64.57}} \\ \hline

\end{tabular}
\end{table*}

\subsubsection{Zero-Shot Cross-Domain Generalization}

In addition to evaluating performance within the training domain, we place special emphasis on assessing cross-domain generalization. To this end, we conduct zero-shot evaluations on 12 external datasets by applying trained models directly, without any fine-tuning. These datasets span diverse tissue types, staining protocols, imaging modalities, and annotation standards (see Figure~\ref{fig:data-overview}), providing a comprehensive assessment of SAMPO’s transferability across varied acquisition protocols and imaging conditions.

\textbf{Comparison with Fine-tuned SOTA Methods.} 
We first benchmarked SAMPO against state-of-the-art (SOTA) supervised methods on four widely-used public datasets: DSB, MoNuSeg, TNBC, and SegPC. As shown in Table~\ref{tab:comparison_sota}, traditional fully supervised methods (e.g., nnU-Net, UN-SAM, PromptNucSeg) generally define the upper bound of performance due to domain-specific training. However, SAMPO achieves remarkable zero-shot performance that rivals these fine-tuned counterparts. Specifically, on the TNBC and MoNuSeg datasets, SAMPO attains Dice scores of 82.13\% and 81.83\%, respectively, surpassing standard U-Net and maintaining a narrow gap with highly specialized models like PromptNucSeg. Notably, compared to vanilla SAM adaptations (e.g., Original SAM-Point) which suffer significant performance drops on pathology data, SAMPO demonstrates superior stability and accuracy. While SAMPO's performance on SegPC (62.45\% F1) is lower than supervised methods, it still significantly outperforms the transferable performance of SAM2 (43.20\% F1) and SAM3 (30.42\% F1), highlighting the difficulty of this specific domain and the robustness of our approach.

\textbf{Robustness Against Foundation Models.}
To further assess robustness, we compared SAMPO directly against the latest generalist foundation models, SAM2 and SAM3, across a diverse array of histological scenarios involving varying nuclear appearances and staining techniques. Table~\ref{tab:comparison_sam_series} summarizes the results across eight diverse benchmarks, including H\&E stained tissues (CoNSeP, CPM15/17, Kumar), fluorescence microscopy, and cryosection data.

\begin{itemize}
    \item \textit{Staining Invariance:} SAMPO demonstrates exceptional generalization across staining modalities. On the Fluorescence dataset, where nuclei appear with significantly different visual characteristics than standard H\&E, SAMPO achieves a Dice score of 90.75\%, exceeding SAM2 (61.98\%) by nearly 30\%. Similarly, on CryoNuSeg, SAMPO maintains a high Dice score of 77.94\% compared to SAM2's 35.78\%.
    \item \textit{Performance in Dense Scenarios:} In datasets characterized by high-density nuclear distribution, such as CPM17 and CoNSeP, generic foundation models frequently fail to capture the nuclear regions correctly, often mistaking crowded clusters for background or non-target tissue. SAMPO demonstrates superior semantic recognition capabilities in these scenarios. On CPM17, it achieves a Dice score of 81.14\%, nearly doubling the performance of SAM2 (42.29\%). This significant margin indicates that SAMPO effectively discerns specific nuclear textures from the complex background even when nuclei are tightly packed, whereas generic models struggle to adapt to such domain-specific density.
    \item \textit{Task Versatility:} Whether for universal segmentation (T1) or category-specific tasks (T2 on CoNSeP), SAMPO consistently outperforms generic SAM baselines, validating that incorporating pathology-specific semantic priors into the promptable framework is essential for trustworthy medical image analysis.
\end{itemize}

\begin{table*}[]
\setlength{\tabcolsep}{3pt} 
\centering
\caption{Zero-shot cross-domain generalization results (Part 1). The SAMPO is trained on PanNuke (10\% data) and evaluated directly on external datasets without fine-tuning. Best results are shown in \textbf{bold}. Manual prompt: MP. $^\dagger$: Reported in \cite{chen2025sam}.}
\label{tab:unsam}
\small
\begin{tabular}{c|c|c|ccc|ccc|ccc|ccc}
\hline
\multirow{2}{*}{Methods} & \multirow{2}{*}{MP}                     & \multirow{2}{*}{Fine-tuned} & \multicolumn{3}{c|}{DSB}                         & \multicolumn{3}{c|}{MoNuSeg}                     & \multicolumn{3}{c|}{TNBC}                        & \multicolumn{3}{c}{SegPC}                        \\ \cline{4-15} 
                         &                                         &                             & Dice           & mIoU           & F1             & Dice           & mIoU           & F1             & Dice           & mIoU           & F1             & Dice           & mIoU           & F1             \\ \hline
U-Net $^\dagger$            & \multirow{6}{*}{\ding{55}} & \checkmark   & 88.16          & 81.42          & 89.49          & 74.06          & 60.25          & 75.57          & 80.64          & 67.62          & 81.00          & 84.89          & 76.20          & 86.75          \\  
Unet++ $^\dagger$           &                                         & \checkmark   & 90.48          & 83.53          & 90.96          & 76.78          & 62.97          & 78.31          & 81.19          & 68.44          & 81.60          & 85.61          & 77.10          & 87.04          \\
Attention-UNet $^\dagger$   &                                         & \checkmark   & 91.38          & 84.76          & 91.79          & 76.89          & 62.92          & 77.45          & 81.25          & 68.59          & 81.84          & 84.25          & 75.91          & 85.75          \\
TransUNet $^\dagger$        &                                         & \checkmark   & 90.76          & 83.73          & 91.08          & 75.08          & 61.41          & 76.55          & 78.30          & 64.37          & 77.65          & 82.91          & 73.85          & 84.53          \\
nnU-Net $^\dagger$          &                                         & \checkmark   & 91.61          & 85.00          & 91.96          & 81.09          & 68.28          & 81.24          & 82.11          & 69.74          & 82.35          & 87.35          & 79.32          & 88.48          \\
UN-SAM $^\dagger$           &                                         & \checkmark   & 93.12          & 87.41          & 93.30          & 84.17          & 72.93          & 84.27          & 85.72          & 75.02          & 85.81          & 89.01          & 82.14          & 89.88          \\ \hline
Original SAM $^\dagger$     & \multirow{8}{*}{\ding{55}} & \checkmark   & 36.49          & 25.83          & 49.04          & 20.28          & 12.62          & 25.25          & 20.44          & 12.53          & 35.80          & 39.88          & 19.17          & 40.40          \\
SAMed $^\dagger$            &                                         & \checkmark   & 82.62          & 75.07          & 82.94          & 65.76          & 51.82          & 65.96          & 75.74          & 61.42          & 75.87          & 77.96          & 67.22          & 79.74          \\
Med-SA $^\dagger$           &                                         & \checkmark   & 82.84          & 75.24          & 83.14          & 64.91          & 51.02          & 65.06          & 77.59          & 62.95          & 78.63          & 77.59          & 66.92          & 78.87          \\
All-in-SAM $^\dagger$       &                                         & \checkmark   & 90.73          & 83.86          & 91.20          & 78.38          & 65.94          & 79.21          & 79.25          & 66.14          & 80.62          & 85.51          & 78.59          & 86.75          \\
H-SAM $^\dagger$            &                                         & \checkmark   & 91.52          & 84.79          & 91.83          & 82.13          & 69.75          & 82.27          & 79.45          & 65.63          & 79.81          & 85.92          & 76.17          & 86.05          \\
CellSAM $^\dagger$          &                                         & \checkmark   & 92.14          & 85.73          & 92.35          & 82.28          & 69.96          & 82.40          & 81.12          & 66.83          & 81.29          & 83.16          & 74.05          & 83.51          \\
SAC $^\dagger$              &                                         & \checkmark   & 91.36          & 87.35          & 91.03          & 82.89          & 69.22          & 82.98          & 81.18          & 68.38          & 81.65          & 87.45          & 79.27          & 88.32          \\
PromptNucSeg $^\dagger$     &                                         & \checkmark   & 92.29          & 86.18          & 92.81          & 82.62          & 69.87          & 79.11          & 84.59          & 73.35          & 84.53          & 87.39          & 80.83          & 87.79          \\ \hline
Original SAM $^\dagger$     & \multirow{5}{*}{Point}                  & \checkmark   & 70.72          & 59.10          & 74.56          & 37.54          & 22.47          & 46.33          & 83.46          & 46.78          & 64.83          & 62.67          & 44.78          & 72.74          \\
SAMed $^\dagger$            &                                         & \checkmark   & 92.39          & 86.27          & 92.66          & 82.26          & 69.94          & 82.43          & 83.19          & 71.23          & 83.33          & 86.87          & 78.68          & 87.90          \\
Med-SA $^\dagger$           &                                         & \checkmark   & 92.52          & 86.45          & 92.80          & 82.47          & 70.24          & 82.57          & 84.16          & 72.68          & 84.32          & 87.39          & 79.54          & 88.54          \\ 
SAM2                       &                & \ding{55}      & 57.98          & 43.90          & 58.24          & 44.51          & 29.81          & 44.85          & 42.81          & 29.44          & 43.50          & 44.34          & 30.43          & 43.20          \\
SAMPO                      &                & \ding{55}      & \textbf{87.04} & \textbf{78.58} & \textbf{86.60} & \textbf{81.83} & \textbf{69.25} & \textbf{82.72} & \textbf{82.13} & \textbf{70.16} & \textbf{83.40} & \textbf{62.67} & \textbf{47.54} & \textbf{62.45} \\ \hline
SAM3                       &  Text              & \ding{55}      & 62.20          & 71.28          & 61.49          & 31.56          & 45.75          & 31.47          & 21.59          & 32.63          & 21.82          & 30.90          & 44.87          & 30.42          \\   \hline
\end{tabular}
\end{table*}

\begin{table*}[!ht]
\centering
\caption{Zero-shot cross-domain generalization results (Part 2). The SAMPO is trained on PanNuke (10\% data) and evaluated directly on external datasets without fine-tuning. Best results are shown in \textbf{bold}.}
\label{tab:sam_series1}
\begin{tabular}{c|cl|cl|cl|cl|cl}
\hline
\multirow{2}{*}{Methods} & \multicolumn{2}{c|}{CoNSeP-T1} & \multicolumn{2}{c|}{CoNSeP-T2}    & \multicolumn{2}{c|}{GlandSegData} & \multicolumn{2}{c|}{Histology} & \multicolumn{2}{c}{Fluorescence}    \\  \cline{2-11}   
                 & \multicolumn{1}{c}{IoU} & Dice & \multicolumn{1}{c}{IoU} & Dice & \multicolumn{1}{c}{IoU}   & Dice & \multicolumn{1}{c}{IoU}   & Dice  & \multicolumn{1}{c}{IoU} & Dice  \\    \hline
SAM2             & 23.29          & 36.63           & 25.36             & 37.91           & 50.27           & 64.78           & 31.85            & 46.91            & 50.05            & 61.98\\
SAM3             & 23.65          & 35.48           & 27.91             & 37.74           & 50.47           & 64.33           & 37.43            & 52.46            & 77.64            & 85.51\\
SAMPO            & \textbf{59.00} & \textbf{73.52}  & \textbf{40.88}    & \textbf{55.46}  & \textbf{61.20}  & \textbf{74.26}  & \textbf{55.16}   & \textbf{70.16}   & \textbf{84.30}   & \textbf{90.75}\\ \hline 
\end{tabular}
\end{table*}

\begin{table*}[!ht]
\centering
\caption{Zero-shot cross-domain generalization results (Part 3). The SAMPO is trained on PanNuke (10\% data) and evaluated directly on external datasets without fine-tuning. Best results are shown in \textbf{bold}. Kumar-Same and Kumar-Diff denote test sets with organ types seen and unseen during training, respectively.}
\label{tab:sam_series2}
\begin{tabular}{c|cl|cl|cl|cl|cl}
\hline
\multirow{2}{*}{Methods} & \multicolumn{2}{c|}{CPM15}   & \multicolumn{2}{c|}{CPM17}    & \multicolumn{2}{c|}{Kumar-same} & \multicolumn{2}{c|}{Kumar-diff} & \multicolumn{2}{c}{CryoNuSeg}   \\ \cline{2-11} 
                         & \multicolumn{1}{c}{IoU} & Dice & \multicolumn{1}{c}{IoU} & Dice & \multicolumn{1}{c}{IoU}   & Dice   & IoU&Dice   & \multicolumn{1}{c}{IoU} & Dice  \\      \hline
SAM2                 & 28.61         & 43.10         & 28.20         & 42.29         & 22.76         & 36.29         & 26.92         &40.77         & 22.24         & 35.78\\
SAM3                 & 43.96         & 56.07         & 45.38         & 57.87         & 22.57         & 34.53         & 37.30         &53.22         & 50.75         & 64.10\\
SAMPO                & \textbf{67.10}& \textbf{79.28}& \textbf{69.27}& \textbf{81.14}& \textbf{64.91}& \textbf{78.45}& \textbf{71.70}&\textbf{83.19}& \textbf{64.71}& \textbf{77.94}\\ \hline
\end{tabular}
\end{table*}

\section{Discussion and Conclusion}

We present SAMPO, which is the first to achieve preference learning on a promptable visual foundation model, used to infer category information from sparse visual prompts, thereby aligning with human intent. We validated the SAMPO framework across three medical tasks, demonstrating the model's significant potential for intent-aware efficient fine-tuning. Notably, the effectiveness of more complex reinforcement learning-based alignment algorithms requires further verification.

\section{Acknowledgements}
This work was supported by the National Key Research and Development Program (2025YFC2423000), National natural science foundation of China (82372096) and AI for Science Foundation of Fudan University (FudanX24AI038).

\section*{Declaration of Competing Interests}

The authors declare that they have no known competing financial interests or personal relationships that could have appeared to influence the work reported in this paper.

\bibliographystyle{model2-names.bst}\biboptions{authoryear}
\bibliography{main}

\end{document}